\PassOptionsToPackage{table}{xcolor}
\documentclass{article}
\usepackage{iclr2026_conference,times}

\usepackage{amsmath,amsfonts,bm}

\def\eqref#1{equation~\ref{#1}}

\def\1{\bm{1}}

\DeclareMathAlphabet{\mathsfit}{\encodingdefault}{\sfdefault}{m}{sl}
\SetMathAlphabet{\mathsfit}{bold}{\encodingdefault}{\sfdefault}{bx}{n}

\usepackage{hyperref}
\hypersetup{colorlinks=true, linkcolor=red!65!black, citecolor=green!45!black, urlcolor=blue!65!black}
\usepackage{url}
\usepackage{amssymb}
\usepackage{array}
\usepackage{booktabs}
\usepackage{tabularx}
\usepackage{multirow}
\usepackage{graphicx}
\usepackage{float}
\usepackage{placeins}
\usepackage{wrapfig}
\usepackage{xcolor}
\usepackage{tikz}
\usepackage{needspace}
\usepackage{listings}
\usetikzlibrary{calc}

\definecolor{solventblue}{RGB}{232,242,246}
\definecolor{proprietarybestgray}{RGB}{238,238,238}
\definecolor{promptframe}{RGB}{31,67,111}
\definecolor{promptback}{RGB}{245,248,253}
\definecolor{reasonframe}{RGB}{43,104,99}
\definecolor{reasonback}{RGB}{244,250,248}
\definecolor{toolframe}{RGB}{166,91,57}
\definecolor{toolback}{RGB}{253,248,244}
\lstdefinestyle{prompt}{
  basicstyle=\ttfamily\scriptsize,
  breaklines=true,
  breakatwhitespace=false,
  columns=fullflexible,
  keepspaces=true,
  showstringspaces=false,
  frame=single,
  framerule=0.4pt,
  rulecolor=\color{black!30},
  backgroundcolor=\color{solventblue},
  xleftmargin=0.5em,
  xrightmargin=0.5em,
  framexleftmargin=0.4em,
  framexrightmargin=0.4em,
  aboveskip=6pt,
  belowskip=6pt,
  captionpos=t
}
\lstdefinestyle{trajectory}{
  basicstyle=\ttfamily\scriptsize,
  breaklines=true,
  breakatwhitespace=false,
  columns=fullflexible,
  keepspaces=true,
  showstringspaces=false,
  frame=single,
  framerule=0.4pt,
  rulecolor=\color{black!25},
  backgroundcolor=\color{black!2},
  xleftmargin=0.5em,
  xrightmargin=0.5em,
  framexleftmargin=0.4em,
  framexrightmargin=0.4em,
  aboveskip=5pt,
  belowskip=7pt,
  literate=
    {→}{{$\rightarrow$}}1
    {α}{{$\alpha$}}1
    {–}{{--}}1
    {₂}{{\textsubscript{2}}}1,
  captionpos=t
}
\lstdefinestyle{tooltrace}{
  style=trajectory,
  rulecolor=\color{blue!35!black},
  backgroundcolor=\color{solventblue}
}
\lstdefinestyle{appendixlisting}{
  basicstyle=\ttfamily\scriptsize,
  breaklines=true,
  breakatwhitespace=false,
  columns=fullflexible,
  keepspaces=true,
  showstringspaces=false,
  frame=single,
  framerule=0.65pt,
  framesep=1.2mm,
  aboveskip=0pt,
  belowskip=7pt,
  xleftmargin=0pt,
  xrightmargin=0pt,
  framexleftmargin=0.8mm,
  framexrightmargin=0.8mm,
  framextopmargin=0.8mm,
  framexbottommargin=0.8mm
}
\lstdefinestyle{appendixtrajectory}{
  style=appendixlisting,
  literate=
    {→}{{$\rightarrow$}}1
    {α}{{$\alpha$}}1
    {–}{{--}}1
    {₂}{{\textsubscript{2}}}1
}
\lstdefinestyle{promptpanel}{
  style=appendixlisting,
  rulecolor=\color{promptframe},
  backgroundcolor=\color{promptback}
}
\lstdefinestyle{reasonpanel}{
  style=appendixtrajectory,
  rulecolor=\color{reasonframe},
  backgroundcolor=\color{reasonback}
}
\lstdefinestyle{toolpanel}{
  style=appendixlisting,
  rulecolor=\color{toolframe},
  backgroundcolor=\color{toolback}
}
\newcommand{\ListingHeader}[2]{%
  \par\noindent
  {\setlength{\fboxsep}{1.2mm}\setlength{\fboxrule}{0.65pt}%
   \fcolorbox{#1}{#1}{%
     \parbox{\dimexpr\linewidth-2\fboxsep-2\fboxrule\relax}{%
       \color{white}\bfseries\small #2}}}%
  \par\nobreak\vspace{-0.65pt}%
}
\newcommand{\PromptListing}[2]{%
  \ListingHeader{promptframe}{#1}%
  \lstinputlisting[style=promptpanel]{#2}%
}
\newcommand{\ReasoningListing}[3][]{%
  \ListingHeader{reasonframe}{#2}%
  \lstinputlisting[style=reasonpanel,#1]{#3}%
}
\newcommand{\ToolListing}[2]{%
  \ListingHeader{toolframe}{#1}%
  \lstinputlisting[style=toolpanel]{#2}%
}
\newcolumntype{C}[1]{>{\centering\arraybackslash}m{#1}}
\newcolumntype{L}[1]{>{\raggedright\arraybackslash}m{#1}}
\newcolumntype{Y}{>{\centering\arraybackslash}X}
\newcommand{\frontierbest}[1]{\cellcolor{proprietarybestgray}#1}
\newcommand{\compactbest}[1]{\textbf{#1}}

\title{Solvent: Long-Horizon Reasoning for Multi-Step Retrosynthesis}

\author{%
Yifei Wang$^{1,3}$, Botian Wang$^{2}$, YiFei Wang$^{2}$, Jiangtao Feng$^{2}$,\\
\textbf{Xinbo Zhang$^{3}$, Yawen Ouyang$^{3}$, Yixin Wu$^{3}$, Lihao Wang$^{3}$,}\\
\textbf{Xiaoqing Zheng$^{1}$, Xuanjing Huang$^{1}$, Hao Zhou$^{2,3}$\thanks{Corresponding author.}}\\[4pt]
$^{1}$Fudan University\quad $^{2}$Generative Symbolic Intelligence Lab (GenSI), Tsinghua University\\
$^{3}$Shanghai Artificial Intelligence Laboratory\\
\texttt{yifeiwang26@m.fudan.edu.cn}, \texttt{zhouhao@air.tsinghua.edu.cn}
}

\iclrfinalcopy
\begin{document}

\maketitle
\lhead{Preprint}

\begin{abstract}
Multi-step retrosynthesis requires recursively transforming a target molecule through successive reaction steps into purchasable building blocks, and recent work has begun to draw on large language model (LLM) reasoning to tackle the task. In its latest evolution, LLM reasoning has entered the long-horizon regime: models pursue objectives across interdependent steps, interact with environments, and adapt to feedback. This shift has markedly advanced LLM capabilities, as illustrated by breakthroughs in AI-driven mathematical research. At this frontier, we introduce \textbf{Solvent}, an 8B LLM that successfully extends long-horizon reasoning to multi-step retrosynthesis. Through a two-round post-training curriculum, Solvent learns to interleave iterative synthesis planning, external availability verification, and adaptive route revision. Experiments show that, following a $2.1\times$ performance boost from RL, Solvent scores 61.86\% across four benchmarks, surpassing mid-scale language models and prior search-based retrosynthesis systems, and rivaling frontier trillion-parameter LLMs.\footnote{Code: \href{https://github.com/alden199/Solvent}{\texttt{github.com/alden199/Solvent}}; model: \href{https://huggingface.co/Alden19910/Solvent}{\texttt{huggingface.co/Alden19910/Solvent}}}
\end{abstract}

\vspace{-10pt}
\begin{figure}[H]
    \centering
    \begin{minipage}{0.88\textwidth}
        \centering
        \includegraphics[width=\linewidth]{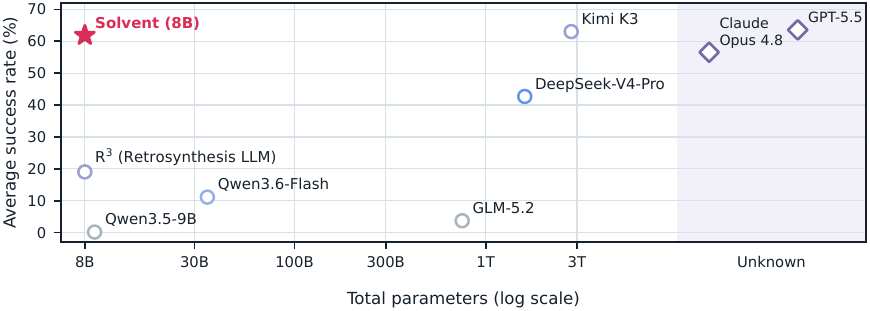}
        \setlength{\abovecaptionskip}{-10pt}
        \caption{At 8B parameters, Solvent substantially outperforms mid-scale LLMs and rivals frontier trillion-parameter models. Proprietary models with undisclosed parameter counts are arbitrarily positioned in the shaded region.}
        \label{fig:model-scale}
    \end{minipage}
\end{figure}

\section{Introduction}
\label{sec:introduction}

Retrosynthetic planning plays a central role in organic synthesis and drug
discovery~\citep{Corey_1991,Zheng2022} by working backward from a complex target molecule to
identify simpler, readily accessible precursors from which the target can be synthesized. This task is
studied at two levels of granularity. Single-step retrosynthesis predicts the immediate reactants
capable of producing a specified product through one chemical transformation~\citep{Jiang_2023}.
Multi-step planning further requires a sequence of interdependent decisions that recursively
decomposes the target into commercially available building
blocks~\citep{Zheng2022,segler-etal-2018-mcts,pmlr-v119-chen20k}.

Traditional AI-based multi-step retrosynthesis casts route planning as an explicit search built on
single-step reaction proposals. Progress has come from improving the local proposal model itself and
from learning value functions that prioritize expansions in search algorithms~\citep{segler-etal-2018-mcts,pmlr-v119-chen20k,pmlr-v139-kim21b,Hong2023}. This decomposition
has made large-scale synthesis planning practical, but it also ties route recovery to the coverage of
the single-step model, incurs substantial search costs as routes deepen, and leaves both the global chemical
analysis of the target and the rationale behind individual bond disconnections implicit in the search
procedure.

Recently, large language models (LLMs) have evolved from single-turn chatbots into long-horizon
agents, demonstrating remarkable capabilities in mathematics, coding, and scientific problem solving
through persistent iteration across reasoning, verifying, and
revising~\citep{dong2026longhorizon,openai2026tenadvances,anthropic2026riemann}. However, this
paradigm remains underexplored in multi-step retrosynthesis, a task naturally suited
to long-horizon reasoning, with route construction involving recursive subproblems, precursor
checking, and feedback-guided revision. Existing approaches
refine LLM-generated routes through external optimization or
search~\citep{pmlr-v267-wang25ag,song-etal-2026-aot}, guide planning over reaction candidates
supplied by single-step
models~\citep{NEURIPS2025_66977686,liao-etal-2026-retrorchestrator,zhu-etal-2026-retroagent}, or
generate a complete route directly in one turn~\citep{wang-etal-2026-r3}.
Across these designs, the LLM's reasoning capacity remains constrained.

We introduce \textbf{Solvent}, an 8B LLM that frames multi-step retrosynthesis
as a long-horizon task, solving it through iterative planning, external verification, and adaptive
revision. Solvent jointly reasons over global strategy and stepwise
disconnections, checks precursor availability against a building-block inventory, and updates
proposed routes accordingly, keeping all retrosynthetic decisions within the model.
Achieving this requires retrosynthetic knowledge and the capacity for continual interaction with the
environment, which we develop in Solvent through a two-round, multistage post-training curriculum.
The first round develops a seed policy that incorporates retrosynthetic knowledge and exhibits a
stable pattern of long-horizon reasoning; the second bootstraps a stronger starting policy,
preparing the model for final RL that boosts its capabilities.
The resulting model shows strategic chemical insight, devising
protecting-group schemes and convergent routes where appropriate.

Across four benchmarks, Solvent achieves leading performance and demonstrates progressive
self-improvement through long-horizon interaction. Among the evaluated LLMs, Solvent reaches an average success
rate of 61.86\%, rivaling the strongest frontier models while surpassing the similarly sized,
retrosynthesis-specialized LLM $R^3$ by 42.83 percentage points. Compared with prior search-based
retrosynthesis systems, Solvent reaches an average success rate of 93.22\% at a search budget of
100, achieving the highest success rates across the budget range. As the interaction horizon
expands, Solvent's average success rate rises from 18.75\% to 61.86\%, demonstrating the benefit
that sustained feedback-based revision provides for Solvent.

Our main contributions are:
\begin{itemize}
    \item \textbf{Framing Multi-Step Retrosynthesis as a Long-Horizon Task.} Solvent addresses
    the problem through iterative planning, external verification, and adaptive revision, keeping
    all retrosynthetic decisions inside the model.

    \item \textbf{Developing Long-Horizon Reasoning through Post-Training.} Our two-round,
    multistage curriculum builds Solvent's domain knowledge and capacity to reason across turns,
    interpret environmental feedback, and make adaptive revisions.

    \item \textbf{Establishing Strong Performance at 8B Parameters.} Solvent rivals frontier
    LLMs, substantially surpasses similarly sized LLMs, and consistently outperforms prior
    search-based retrosynthesis systems across the search-budget range.
\end{itemize}

\section{Related Work}
\label{sec:related-work}

\subsection{Long-Horizon Tasks and Agents}

Long-horizon tasks are characterized by dependencies among decisions: reaching a goal requires
multiple steps, with earlier results shaping later choices~\citep{dong2026longhorizon}.
Recent work examines such tasks in software
engineering~\citep{NEURIPS2025_85069585}, web
navigation~\citep{pmlr-v267-erdogan25a}, and realistic workplace
assistance~\citep{NEURIPS2025_0d744742}.

Long-horizon agents are designed to solve such tasks through sustained interaction with an
environment~\citep{dong2026longhorizon}. Recent work has studied this capability across code
generation, tool-assisted reasoning, and software engineering, using execution or task feedback to
guide subsequent steps~\citep{pmlr-v267-gehring25a,feng-etal-2026-retool,openai-2025-codex}.

\subsection{Large Language Models for Multi-Step Retrosynthesis}

Recent studies have begun to harness LLMs for multi-step retrosynthetic planning. One line of work
takes a search-based approach. These methods either integrate LLM-generated routes or
disconnections into external search procedures, or use LLMs to coordinate search over precursor
candidates supplied by external single-step models. LLM-Syn-Planner improves a population of
complete routes through evolutionary search~\citep{pmlr-v267-wang25ag}. AOT* maps complete
pathways proposed by an LLM into an AND--OR tree and applies explicit tree search to select unsolved
intermediates~\citep{song-etal-2026-aot}. DeepRetro combines an external recursive search procedure
driven by a template-based or MCTS-based retrosynthesis solver with an LLM that proposes and validates
single-step disconnections~\citep{Sathyanarayana_2026}. Retro-R1 uses an LLM to select among precursor sets proposed by an
external single-step model~\citep{NEURIPS2025_66977686}; RetrOrchestrator uses an LLM to choose
the molecule to expand, the single-step model, and the requested number of predictions~\citep{liao-etal-2026-retrorchestrator}; and RetroAgent uses an LLM to select graph expansions
while an external environment maintains the AND--OR graph and supplies single-step reaction
candidates~\citep{zhu-etal-2026-retroagent}. In these methods, route exploration or precursor
generation is handled outside the LLM.

A separate approach lets the LLM generate a complete route directly. $R^3$ produces a complete
retrosynthetic route in a single open-loop turn~\citep{wang-etal-2026-r3}.

Solvent instead frames multi-step retrosynthesis as a long-horizon task in which the LLM carries out
global planning, stepwise decomposition, inventory checking, and feedback-guided revision over
multiple turns. More related work is discussed in Appendix~\ref{app:related-work}.

\section{Preliminaries}
\label{sec:preliminaries}

\paragraph{Multi-step retrosynthesis.}
Let $\mathcal{M}$ denote the molecular space and $\mathcal{B}\subset\mathcal{M}$ the set of
available building blocks. Given a target molecule $x\in\mathcal{M}\setminus\mathcal{B}$,
multi-step retrosynthesis seeks a rooted, acyclic reaction tree
$\mathcal{T}=(\mathcal{V},\mathcal{E})$ whose root is $x$~\citep{pmlr-v119-chen20k,NEURIPS2025_66977686,wang-etal-2026-r3}.
Each internal molecule $m\in\mathcal{V}$ is decomposed by a reaction
$e:m\Rightarrow\{m_1,\ldots,m_k\}$, and the leaves $\mathcal{L}(\mathcal{T})$ are the terminal
precursors. For simplicity, we call a route \emph{stock-complete} if
\begin{equation}
    \operatorname{root}(\mathcal{T})=x,
    \qquad
    \mathcal{L}(\mathcal{T})\subseteq\mathcal{B}.
\end{equation}
A stock-complete solution is then further assessed for the plausibility of each reaction.
Further preliminaries on agentic reinforcement learning is provided in Appendix~\ref{app:agentic-rl}.

\section{Methodology}
\label{sec:methodology}

We frame multi-step retrosynthesis as a long-horizon task. Over multiple turns, Solvent iteratively
plans a route, checks precursor availability against a building-block inventory, and revises the route
in response to inventory feedback. Unlocking the model's full potential through agentic RL requires
a strong starting point with clear, stable learning signals,
motivating our two-round training pipeline: (1) a first round that develops a seed policy carrying retrosynthetic knowledge and a
stable pattern of long-horizon reasoning, and (2) a second round that uses rejection-sampling
fine-tuning to bootstrap a stronger starting policy and finally applies RL to boost long-horizon
reasoning.

\begin{figure}[!t]
    \centering
    \includegraphics[width=\textwidth]{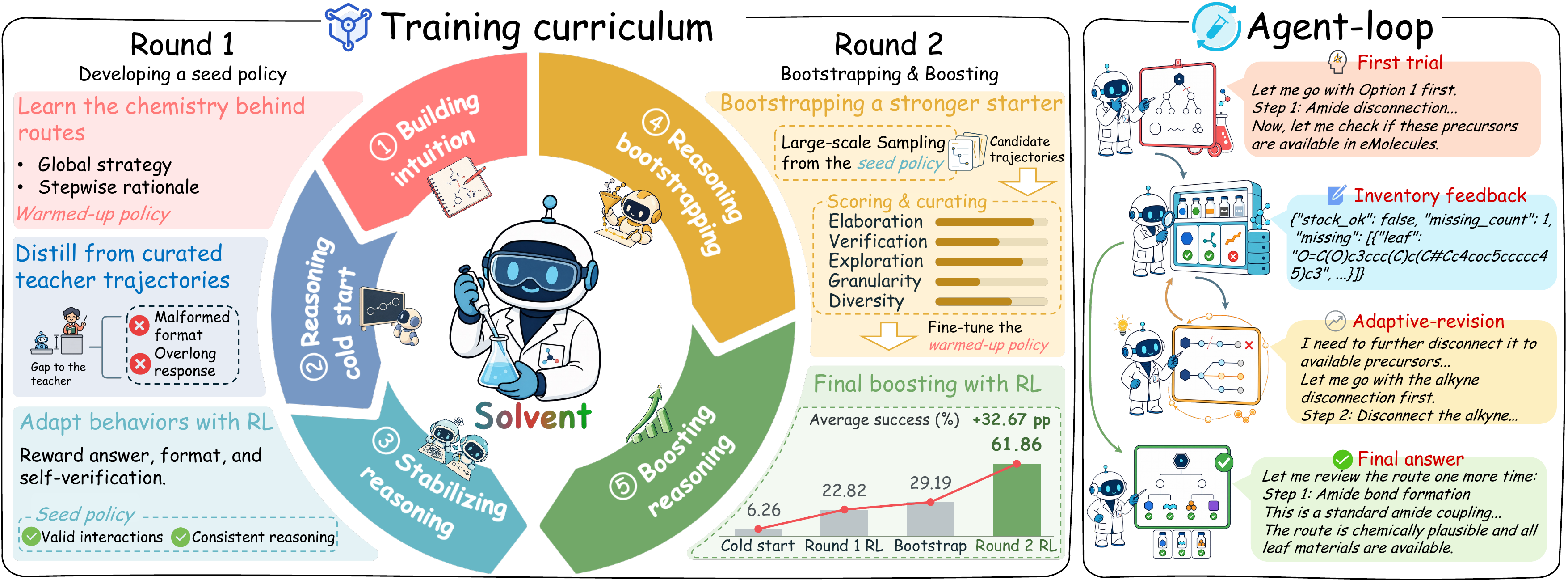}
    \setlength{\abovecaptionskip}{0pt}
    \caption{\textbf{Overview of Solvent.}
    \textbf{Left:} Two rounds of post-training prepare Solvent for effective RL.
    Round 1 develops a seed policy with retrosynthetic knowledge and stable reasoning.
    Round 2 bootstraps a stronger starting policy using the seed policy,
    enabling final-stage RL to substantially boost performance.
    \textbf{Right:} Solvent performs long-horizon reasoning through iterative planning,
    external verification, and adaptive revision.}
    \label{fig:solvent-overview}
\end{figure}

\subsection{Developing a Seed Policy for Long-Horizon Reasoning}
\label{sec:initialization}

We first equip the backbone with retrosynthetic knowledge, then cold-start long-horizon reasoning over multiple turns, and finally stabilize this behavior with RL.

\paragraph{Building retrosynthetic intuition.}
We use a teacher model to distill retrosynthetic knowledge from reference routes into structured
analyses, covering target-level strategy and every
reaction step. We retain analyses that faithfully preserve the original
reaction routes. Throughout post-training, we apply a decontamination procedure to training
corpora to remove overlap with evaluation benchmarks. The retained data remain structurally diverse
and well separated from  benchmark targets at both fingerprint and scaffold levels (Appendix~\ref{app:benchmarks}).
We fine-tune the retrosynthesis-pretrained backbone on the corpus, yielding the
warmed-up policy $\pi_{\mathrm{warmed}}$ with retrosynthetic intuition.

\phantomsection\label{para:teacher-cold-start}
\textbf{Teacher-guided cold start for long-horizon reasoning.}
We formulate multi-step retrosynthesis as a long-horizon reasoning task. Given a target molecule, the model develops a global route strategy, makes stepwise disconnection decisions, checks proposed precursors against a building-block inventory, interprets availability feedback, and adapts subsequent decisions over multiple turns. For a target molecule $x$, let $q(x)$ denote the task prompt and
$h_t=((r_i,a_i,o_i))_{i<t}$ the history of reasoning $r_i$, actions $a_i$, and inventory
observations $o_i$ from preceding turns. At turn $t$, the policy samples:
\begin{equation}
    (r_t,a_t)\sim\pi_\theta\!\left(\cdot\mid q(x),h_t\right),
    \label{eq:agentic-retrosynthesis}
\end{equation}
where $r_t$ denotes the reasoning content and $a_t$ either queries the inventory with a
proposed route or returns the final route. The observation $o_i$ reports the availability of
the submitted route's terminal precursors. We sample multi-turn trajectories from a teacher model,
retaining only those that yield valid solutions, remain consistent with the recorded inventory
interactions, and pass the chemical-validity assessment. After strictly removing overlaps with the evaluation
benchmarks, we fine-tune $\pi_{\mathrm{warmed}}$ on the curated trajectories, establishing an initial policy for iterative route construction and revision; data-construction and training details appear in Appendices~\ref{app:teacher-cold-start} and~\ref{app:coldstart-training}.

\paragraph{Consolidating long-horizon reasoning behavior with RL.}\label{para:first-round-rl}
Teacher-guided fine-tuning establishes a basis for long-horizon reasoning. However, the capability gap between the teacher model and the student policy can lead to unstable behavior, including truncated responses and malformed inventory queries. To close this gap, we optimize the policy with Group Sequence Policy Optimization
(GSPO)~\citep{zheng-etal-2025-gspo}, consolidating its own successful behaviors into stable
reasoning patterns. We draw structurally challenging targets with broad molecular diversity from
USPTO-full~\citep{NEURIPS2019_0d2b2061}, exclude all instances that overlap with the evaluation
benchmarks, and apply the following trajectory-level reward:
\begin{equation}
    R_i
    =0.6r_i^{\mathrm{sol}}
    +0.2r_i^{\mathrm{verify}}
    +0.2r_i^{\mathrm{format}}.
    \label{eq:seed-reward}
\end{equation}
The solution term $r_i^{\mathrm{sol}}$ is the binary output of the fixed route verifier used consistently
to define task success throughout training and evaluation. It accepts only nontrivial routes that are
structurally valid, stock-complete, and satisfy the chemical-validity criteria. The other terms reward
inventory verification and format adherence.
We denote the resulting policy by $\pi_{\mathrm{seed}}$ and use it to generate trajectories for
second-round bootstrapping; details of target selection and RL training are provided in
Appendices~\ref{app:first-round-rl-targets} and~\ref{app:first-round-rl-training}.

\subsection{Bootstrapping and Boosting Long-Horizon Reasoning}
\label{sec:reasoning-bootstrap}

We then bootstrap a stronger starting policy through rejection-sampling fine-tuning~\citep{yuan2023scalingrelationshiplearningmathematical}, and finally apply RL to elicit a more
refined mode of long-horizon reasoning.

\subsubsection{Reasoning Bootstrapping}
\label{sec:reasoning-bootstrapping}

Effective RL benefits from an initialization that balances task performance with exploratory
capacity~\citep{li-etal-2026-lightweight-sft}, and we obtain such a policy through reasoning
bootstrapping~\citep{NEURIPS2022_639a9a17}: we sample trajectories from $\pi_{\mathrm{seed}}$ and
fine-tune $\pi_{\mathrm{warmed}}$ on a curated subset of high-quality, diverse trajectories.

\paragraph{Sampling candidate trajectories.}
We assemble a structurally diverse set of challenging targets from
USPTO-full~\citep{NEURIPS2019_0d2b2061}. After strictly removing overlaps with the evaluation
benchmarks, we use $\pi_{\mathrm{seed}}$ to sample multiple trajectories per target, forming the
candidate pool.

\paragraph{Curating for reasoning quality and diversity.}
We first retain only those trajectories that yield a stock-complete route, remain fully consistent
with the recorded inventory interactions, and pass the chemical-validity assessment. Following
LIMO~\citep{ye-etal-2025-limo}, we then score the remaining trajectories along four complementary
axes: reasoning elaboration, self-verification, exploration, and adaptive granularity. These
individual scores are further combined with a reasoning-style diversity term and an $n$-gram
repetition penalty, and we retain only the single highest-scoring trajectory for each canonical
target. Full filtering and scoring details are provided in Appendix~\ref{app:reasoning-bootstrap-data}.

\paragraph{Balancing performance and exploration.}
We fine-tune $\pi_{\mathrm{warmed}}$ on the curated trajectories, carefully tracking validation
success and held-out entropy throughout training in order to detect any premature loss of
exploration. We find that the first-epoch checkpoint offers a favorable
exploration–exploitation trade-off: it substantially improves retrosynthesis performance over
$\pi_{\mathrm{seed}}$ while still avoiding the sharp entropy decline observed with further training
(Appendix~\ref{app:bootstrap-depth-ablation}). The resulting policy, $\pi_{\mathrm{boot}}$, subsequently serves as
the initialization for the final stage of RL.

\subsubsection{Boosting Long-Horizon Reasoning with RL}
\label{sec:final-rl}

Finally, we optimize $\pi_{\mathrm{boot}}$ with GSPO~\citep{zheng-etal-2025-gspo} to further enhance
the policy's long-horizon reasoning. For final-stage RL, we curate a more challenging
and structurally diverse target set
from USPTO-full~\citep{NEURIPS2019_0d2b2061}, with all overlaps with benchmarks removed. To prioritize
route success, we gate $r_i^{\mathrm{verify}}$ and $r_i^{\mathrm{format}}$ on $r_i^{\mathrm{sol}}$:
\begin{equation}
    \begin{aligned}
        \bar r_i^k & =
        \begin{cases}
            r_i^k, & r_i^{\mathrm{sol}}=1,\\
            0, & \text{otherwise},
        \end{cases}
        \quad k\in\{\mathrm{verify},\mathrm{format}\},\\
        R_i^{\mathrm{final}} & =0.8r_i^{\mathrm{sol}}
        +0.1\bar r_i^{\mathrm{verify}}
        +0.1\bar r_i^{\mathrm{format}}.
    \end{aligned}
    \label{eq:final-reward}
\end{equation}
The reward components retain the definitions from Eq.~\ref{eq:seed-reward}, and $\bar r_i^k$
denotes the gated form of the two auxiliary terms. By scoring complete trajectories, the objective
assigns credit to route construction, inventory interaction, and feedback-guided revision when they
jointly produce a successful route. The complete RL configuration is
detailed in Appendix~\ref{app:final-rl-training}.

\section{Experiments}
\label{sec:experiments}

Solvent achieves leading results across four widely used multi-step retrosynthesis benchmarks. Among
the evaluated LLMs, Solvent rivals frontier models while substantially outperforming similarly-sized ones.
Against prior search-based retrosynthesis systems, Solvent records the highest success rates across
the search-budget range. Stage-wise
analyses trace how post-training progressively unlocks Solvent's long-horizon reasoning.
Evaluation settings, additional ablations, and representative cases appear in
Appendices~\ref{app:experimental-settings}, \ref{app:additional-ablations}, and~\ref{app:trajectories}.

\subsection{Experimental Setup}
\label{sec:exp-setup}

\paragraph{Benchmarks.}
We use four widely adopted multi-step retrosynthesis benchmarks. USPTO-190 contains 190
patent-derived targets~\citep{pmlr-v119-chen20k}. ChEMBL-1000 comprises 1,000 structurally complex
compounds sampled from the ChEMBL bioactivity database~\citep{pmlr-v202-liu23as,gaulton2012chembl}.
Pistachio Reachable contains 150 targets with reference routes averaging 5.5 reactions and a mean
depth of 5.4, whereas Pistachio Hard contains 100 targets with longer routes averaging 7.5 reactions
and a mean depth of 7.2~\citep{NEURIPS2024_cd091a4d}. Following standard practice, we use
the building-block set from eMolecules as the inventory of commercially available compounds for
route evaluation and precursor-availability feedback~\citep{pmlr-v119-chen20k,NEURIPS2024_cd091a4d}.

\paragraph{Evaluation.}
We evaluate Solvent under the route-level evaluation protocols established in prior work~\citep{pmlr-v119-chen20k,NEURIPS2024_cd091a4d,wang-etal-2026-r3,song-etal-2026-aot}. Success requires
a valid route tree rooted at the target, all leaves in stock, and every reaction to pass the chemical-validity
check combining plausibility with atom-source accounting (Appendix~\ref{app:benchmarks}).

\paragraph{Baselines.}
We first compare Solvent with proprietary and open-weight LLMs, including frontier models such as
Claude Opus 4.8~\citep{claude-opus-4-8-2026} and Kimi K3~\citep{kimi-k3-2026} and mid-scale models
such as Qwen3.6-Flash~\citep{qwen3-6-2026} and the 8B retrosynthesis model $R^3$~\citep{wang-etal-2026-r3}.
We further compare Solvent with prior search-based retrosynthesis systems on USPTO-190,
ChEMBL-1000, and the Pistachio benchmarks~\citep{pmlr-v119-chen20k,pmlr-v139-kim21b,
Hong2023,pmlr-v202-liu23as,NEURIPS2025_66977686,pmlr-v267-wang25ag,
song-etal-2026-aot,Zhong_2022,segler-etal-2018-mcts}.
Details are provided in Appendix~\ref{app:baseline-budget}.

\paragraph{Inference protocol.}
We select Solvent's inference configuration on a held-out set before test evaluation (Appendices~\ref{app:interaction-turn-selection}
and~\ref{app:inference-setup}). For comparison with LLMs, all models use a common 32k-token output limit and up to five interaction
turns, with each model evaluated under its optimal inference configuration. Comparisons
with prior search-based retrosynthesis systems use aligned search budgets, with the protocol detailed
in Appendix~\ref{app:baseline-budget}.

\begin{table*}[!t]
\centering
\small
\setlength{\tabcolsep}{2.3pt}
\renewcommand{\arraystretch}{1.12}
\renewcommand{\tabularxcolumn}[1]{m{#1}}
\caption{\textbf{At 8B parameters, Solvent remains competitive with frontier LLMs across four multi-step retrosynthesis benchmarks.} We report parameter counts, benchmark-specific success rates, and their average. To enable cost-controlled, target-matched evaluation, selected models are evaluated on a shared 400-target subset and marked with $\dagger$. Em dashes denote undisclosed parameter counts. Gray shading and boldface mark the best frontier and mid-scale results, respectively.}
\label{tab:llm-baselines}
\begin{tabularx}{\textwidth}{L{3.0cm}C{1.25cm}*{4}{Y}C{2.1cm}}
\toprule
Method &
Params. &
\shortstack[c]{Pistachio\\Hard} &
\shortstack[c]{Pistachio\\Reachable} &
USPTO-190 &
\shortstack[c]{ChEMBL-\\1000} &
Average (\%) \\
\midrule
\multicolumn{7}{l}{\textbf{Frontier LLMs (proprietary)}} \\
\midrule
Claude Opus 4.8$^\dagger$ & \textemdash & 38.57 & 77.50 & \frontierbest{54.44} & 55.63 & 56.54 \\
Gemini 3.5 Flash$^\dagger$ & \textemdash & 42.86 & 72.50 & 52.22 & \frontierbest{65.63} & 58.30 \\
Grok 4.5$^\dagger$ & \textemdash & 52.86 & 71.25 & 50.00 & 62.50 & 59.15 \\
GPT-5.5$^\dagger$ & \textemdash & \frontierbest{58.57} & \frontierbest{82.50} & \frontierbest{54.44} & 58.75 & \frontierbest{63.57} \\
\midrule
\multicolumn{7}{l}{\textbf{Frontier LLMs (open-weight)}} \\
\midrule
GLM-5.2$^\dagger$ & $753\,\mathrm{B}$ & 1.43 & 6.25 & 2.22 & 5.00 & 3.73 \\
Kimi K2.5$^\dagger$ & $1\,\mathrm{T}$ & 27.14 & 36.25 & 25.56 & 19.38 & 27.08 \\
DeepSeek-V4-Pro & $1.6\,\mathrm{T}$ & 38.00 & 54.00 & 36.32 & 42.50 & 42.70 \\
Kimi K3$^\dagger$ & $2.8\,\mathrm{T}$ & 54.29 & 80.00 & 52.22 & \frontierbest{65.63} & 63.03 \\
\specialrule{\lightrulewidth}{\aboverulesep}{0.6pt}
\specialrule{\lightrulewidth}{0pt}{\belowrulesep}
\multicolumn{7}{l}{\textbf{Mid-scale LLMs}} \\
\midrule
Qwen3-8B & $8\,\mathrm{B}$ & 0.00 & 0.00 & 0.53 & 0.10 & 0.16 \\
Qwen3.5-9B & $9\,\mathrm{B}$ & 0.00 & 0.67 & 0.00 & 0.00 & 0.17 \\
Qwen3.6-Flash$^\dagger$ & $35\,\mathrm{B}$ & 10.00 & 22.50 & 8.89 & 3.13 & 11.13 \\
$R^3$ & $8\,\mathrm{B}$ & 19.00 & 26.67 & 13.16 & 17.30 & 19.03 \\
\rowcolor{solventblue}
Solvent & $8\,\mathrm{B}$ & \compactbest{62.00} & \compactbest{78.00} & \compactbest{48.95} & \compactbest{58.50} & \compactbest{61.86} \\
\bottomrule
\end{tabularx}
\end{table*}

\subsection{Main Results}
\label{sec:main-results}

We compare Solvent with both LLMs and prior search-based retrosynthesis systems,
demonstrating Solvent's strong performance on multi-step retrosynthesis. Among the evaluated
LLMs, Solvent performs on par with frontier models and substantially outperforms similarly sized
models across the four benchmarks. Against prior search-based retrosynthesis systems, Solvent
achieves the highest success rates across the search-budget range.

\begin{table*}[!t]
\centering
\caption{\textbf{Comparison with prior search-based retrosynthesis systems across search budgets.}
Here, $N$ tracks how much explicit reaction-level construction each method performs before finding a
successful route, with the cross-method accounting defined in Appendix~\ref{app:baseline-budget}.
Bold and underlined values denote the best and second-best results, respectively.}
\label{tab:controlled-budget}
\footnotesize
\setlength{\tabcolsep}{3pt}
\setlength{\arrayrulewidth}{0.3pt}
\renewcommand{\arraystretch}{1.04}

\begin{tabularx}{\textwidth}{L{3.15cm}*{6}{Y}}
\toprule
\multicolumn{7}{l}{\textbf{Performance on USPTO-190}} \\
\midrule
Method & $N=50$ & $N=100$ & $N=200$ & $N=300$ & $N=400$ & $N=500$ \\
\midrule
Greedy DFS & 19.47 & 19.47 & 19.47 & 19.47 & 19.47 & 19.47 \\
retro* & 38.95 & 50.00 & 68.95 & 74.74 & 77.89 & 79.47 \\
retro*-0 & 27.37 & 37.89 & 55.79 & 62.63 & 72.11 & 74.21 \\
retro*+ & 55.79 & 68.42 & 79.47 & 83.16 & 84.21 & 84.21 \\
retro*+-0 & 47.89 & 61.05 & 76.84 & 82.11 & 85.79 & 86.32 \\
EG-MCTS & 58.42 & 64.74 & 69.47 & 75.26 & 78.42 & 81.05 \\
PDVN & 64.74 & 72.11 & \underline{82.63} & \underline{87.89} & \underline{91.58} & \underline{92.11} \\
RETRO-R1 & \underline{73.21} & \underline{77.21} & 82.58 & 84.47 & 85.89 & 86.95 \\
\midrule
\rowcolor{solventblue}
Solvent & \textbf{82.11} & \textbf{87.36} & \textbf{88.95} & \textbf{94.21} & \textbf{96.84} & \textbf{97.37} \\
\specialrule{\heavyrulewidth}{\aboverulesep}{\belowrulesep}
\multicolumn{7}{l}{\textbf{Performance on ChEMBL-1000}} \\
\midrule
Method & $N=50$ & $N=100$ & $N=200$ & $N=300$ & $N=400$ & $N=500$ \\
\midrule
Greedy DFS & 38.10 & 38.10 & 38.10 & 38.10 & 38.10 & 38.10 \\
retro* & 65.60 & 69.00 & 71.60 & 73.20 & 74.00 & 74.40 \\
retro*-0 & 64.60 & 67.20 & 70.30 & 71.80 & 72.70 & 73.40 \\
retro*+ & 70.50 & 73.90 & 76.20 & 78.00 & 78.80 & 79.40 \\
retro*+-0 & 69.40 & 73.60 & 75.60 & 76.70 & 77.60 & 78.50 \\
EG-MCTS & 71.20 & 74.20 & 76.90 & 78.30 & 79.00 & 79.30 \\
PDVN & 73.90 & 76.40 & 78.10 & 79.60 & 80.30 & 80.60 \\
RETRO-R1 & \underline{77.88} & \underline{80.21} & \underline{81.79} & \underline{82.62} & \underline{83.28} & \underline{83.72} \\
\midrule
\rowcolor{solventblue}
Solvent & \textbf{87.10} & \textbf{93.20} & \textbf{95.90} & \textbf{96.70} & \textbf{97.30} & \textbf{98.20} \\
\specialrule{\heavyrulewidth}{\aboverulesep}{\belowrulesep}
\multicolumn{7}{l}{\textbf{Performance on Pistachio Reachable \& Hard}} \\
\midrule
\multirow{2}{*}{Method} &
\multicolumn{3}{c}{\textbf{Pistachio Reachable}} &
\multicolumn{3}{c}{\textbf{Pistachio Hard}} \\
\cmidrule(lr){2-4}\cmidrule(lr){5-7}
& $N=100$ & $N=300$ & $N=500$ &
$N=100$ & $N=300$ & $N=500$ \\
\midrule
LLM (MCTS) & 12.70 & 17.30 & 20.70 & 0.00 & 4.00 & 5.00 \\
LLM (Retro*) & 14.70 & 19.30 & 13.30 & 0.00 & 2.00 & 5.00 \\
LLM-Syn-Planner & 93.30 & 98.00 & 98.00 & 72.00 & 86.00 & 87.00 \\
AOT* & 98.70 & \underline{99.30} & \underline{99.30} & \underline{86.00} & \underline{89.00} & \underline{93.00} \\
RootAligned (MCTS) & \underline{99.30} & \underline{99.30} & \underline{99.30} & 83.00 & 85.00 & 85.00 \\
RootAligned (Retro*) & 98.70 & 98.70 & 98.70 & 78.00 & 82.00 & 82.00 \\
\midrule
\rowcolor{solventblue}
Solvent & \textbf{99.33} & \textbf{100.00} & \textbf{100.00} &
\textbf{93.00} & \textbf{98.00} & \textbf{98.00} \\
\bottomrule
\end{tabularx}
\end{table*}

\paragraph{\textbf{At 8B parameters, Solvent performs competitively with frontier LLMs on multi-step retrosynthesis.}}
Table~\ref{tab:llm-baselines} compares Solvent with leading proprietary and open-weight LLMs. Given
limited evaluation budgets, we follow established practice by evaluating selected models on a common
subset~\citep{openai-gpt5-2025,anthropic-claude37-system-card-2025}, with details provided in
Appendix~\ref{app:baseline-budget}.
Solvent averages 61.86\% on the full test sets, competing with frontier LLMs while substantially outperforming similarly sized open-weight models. For a direct comparison, we also report the results for
all evaluated LLMs on the shared subset (Appendix~\ref{app:baseline-budget}). Appendix~\ref{app:case-feedback-revision}
shows how Solvent revises routes using inventory feedback, while
Appendix~\ref{app:orthogonal-route-assessments} reports further assessments of route quality.

\paragraph{\textbf{Solvent leads prior search-based retrosynthesis systems across search budgets.}}
Solvent achieves the highest success rates on all four benchmarks throughout the budget
range (Table~\ref{tab:controlled-budget}). We leverage larger search budgets through parallel sampling,
enabling Solvent to explore a broader range of retrosynthetic strategies; Appendix~\ref{app:case-parallel-sampling}
compares routes produced in independent trials.

\subsection{Ablation Study}
\label{sec:ablations}

\paragraph{\textbf{Successive post-training stages progressively unlock Solvent's long-horizon
reasoning, with final-stage RL accounting for the largest gain.}}
Table~\ref{tab:training-stages} compares the initial backbone with cumulative stage-wise checkpoints
across the four benchmarks. Average success increases at every stage, from 0.18\% to 61.86\%.

\paragraph{\textbf{The first training round develops a seed policy that exhibits a stable pattern of long-horizon reasoning.}}
After teacher-guided cold start, average success increases from 0.18\% to 6.26\%, while truncated
responses and malformed inventory queries remain frequent. Round 1 RL substantially stabilizes the
model's response patterns, raising success to 22.82\%. Appendix~\ref{app:case-round1-evolution}
compares the model's reasoning on the same target before and after Round 1 RL.

\paragraph{\textbf{Bootstrapping consolidates reasoning patterns and preserves exploration, setting the
stage for capability emergence in final-stage RL.}}

\begin{table*}[!t]
\centering
\caption{\textbf{Stage-wise performance across the two-round post-training curriculum.}
We report performance for four training stages described in
\hyperref[sec:initialization]{Sec.~4.1} and
\hyperref[sec:reasoning-bootstrap]{Sec.~4.2}. Average success improves across the reported
stages, with the largest gain occurring after final-stage RL.}
\label{tab:training-stages}
\footnotesize
\setlength{\tabcolsep}{3pt}
\renewcommand{\arraystretch}{1.12}
\renewcommand{\tabularxcolumn}[1]{m{#1}}
\begin{tabularx}{\textwidth}{L{1.35cm}L{3.65cm}*{5}{Y}}
\toprule
\multicolumn{2}{l}{Training stage} &
\shortstack[c]{Pistachio\\Hard} &
\shortstack[c]{Pistachio\\Reachable} &
\shortstack[c]{USPTO-\\190} &
\shortstack[c]{ChEMBL-\\1000} &
Average \\
\midrule
\multicolumn{2}{l}{Initial Backbone}
& 0.00 & 0.00 & 0.53 & 0.20 & 0.18 \\
\midrule
\multirow[c]{2}{*}{\textbf{Round 1}} & Teacher-guided cold start
 & 2.00 & 12.67 & 3.16 & 7.20 & 6.26 \\
& First-round RL
 & 19.00 & 40.00 & 9.47 & 22.80 & 22.82 \\
\specialrule{\heavyrulewidth}{\aboverulesep}{\belowrulesep}
\multirow[c]{2}{*}{\textbf{Round 2}} & Reasoning bootstrapping
 & 23.00 & 41.33 & 22.63 & 29.80 & 29.19 \\
& \cellcolor{solventblue}Final-stage RL
& \cellcolor{solventblue}\textbf{62.00} & \cellcolor{solventblue}\textbf{78.00}
& \cellcolor{solventblue}\textbf{48.95} & \cellcolor{solventblue}\textbf{58.50}
& \cellcolor{solventblue}\textbf{61.86} \\
\bottomrule
\end{tabularx}
\end{table*}

Round 2 bootstrapping produces a stronger starting point for final-stage RL, reaching 29.19\%
average success after fine-tuning on curated trajectories. Appendix~\ref{app:two-round-curriculum-ablation}
further examines the role of bootstrapping. Final-stage RL yields the largest improvement, raising
success from 29.19\% to 61.86\%.

\section{Analysis}
\label{sec:analysis}

We analyze the emergence of Solvent's long-horizon capabilities throughout post-training. We first
examine how post-training enhances Solvent's ability to revise routes adaptively. We then examine how
final-stage RL progressively improves Solvent's long-horizon reasoning.

\subsection{The Capacity for Adaptive Revision Emerges across Post-Training}
\label{sec:analysis-depth}

\begin{figure}[!t]
\centering
\includegraphics[width=\textwidth]{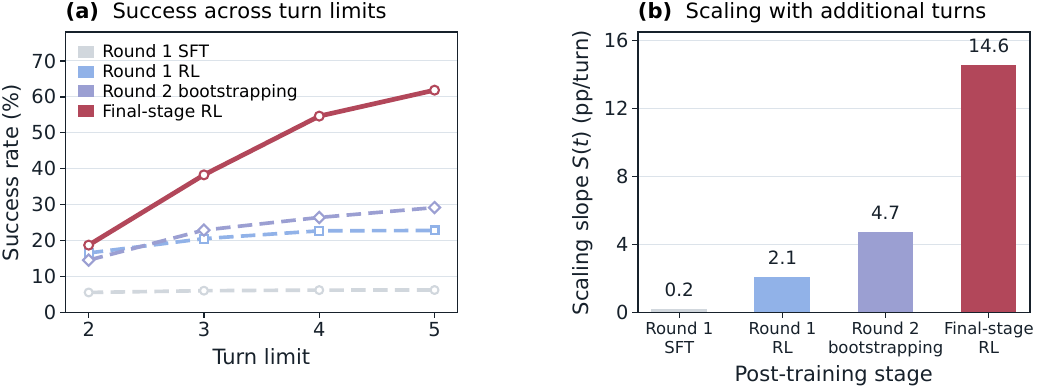}
\setlength{\abovecaptionskip}{0pt}
\caption{\textbf{The benefit of additional interaction turns increases throughout post-training.}
(a) Average success rate under turn limits of 2--5.
(b) Success gain per additional turn across stages.}
\label{fig:checkpoint-feedback-scaling}
\end{figure}

\paragraph{\textbf{Solvent progressively learns to use feedback for revision during post-training.}}
To examine this, we evaluate Solvent across post-training stages with turn limits of two to five on all
four benchmarks (Figure~\ref{fig:checkpoint-feedback-scaling}(a)). Figure~\ref{fig:checkpoint-feedback-scaling}(b)
reports the success gain from an additional turn at each post-training stage as the ordinary least-squares slope
$S(t)$, in percentage points per turn (pp/turn). The results show that, over the course of post-training, Solvent learns to improve its proposals
through environmental feedback. Appendix~\ref{app:case-convergent-route} provides a case study of self-revision.

\subsection{How Final-Stage RL Improves Long-Horizon Route Construction}
\label{sec:analysis-rl}

\begin{figure}[!ht]
\centering
\includegraphics[width=\textwidth]{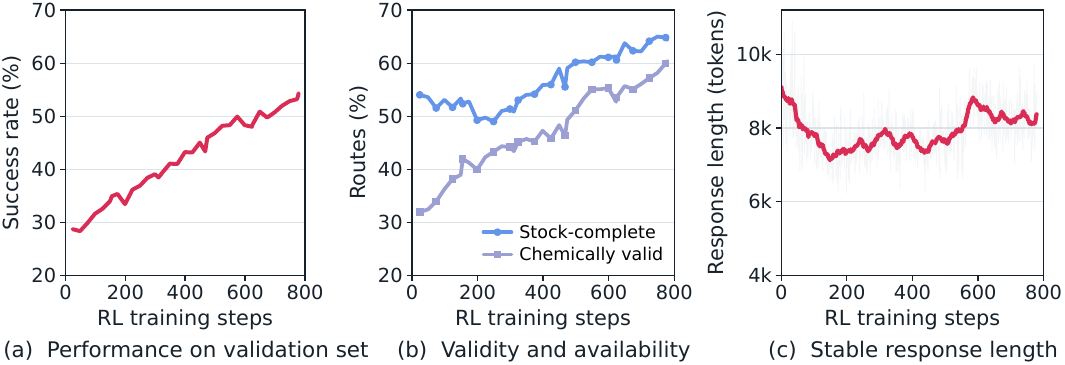}
\setlength{\abovecaptionskip}{0pt}
\caption{\textbf{Final-stage RL successively improves chemical validity and stock completion.}
(a) Performance on validation set during final-stage RL.
(b) Evolution of chemical validity and stock completion.
(c) Mean response length during training.}
\label{fig:rl-route-construction}
\end{figure}

\paragraph{\textbf{Final-stage RL progressively improves chemical validity and stock completion during
long-horizon route construction.}}
Validation success rises throughout final-stage RL (Figure~\ref{fig:rl-route-construction}(a)).
Chemical validity improves first, followed by gains in stock completion
(Figure~\ref{fig:rl-route-construction}(b)), indicating that final-stage RL first improves individual
disconnection decisions and then integrates them into stock-complete routes. The mean response length
remains stable (Figure~\ref{fig:rl-route-construction}(c)).
After final-stage RL, Solvent exhibits the capacity for strategic synthesis
planning, including protecting-group use and convergent route design
(Appendix~\ref{app:case-feedback-revision}).
Appendix~\ref{app:case-final-rl-evolution} presents a case study comparing
chemical strategies before and after final-stage RL.
Further training dynamics appear in Appendix~\ref{app:rl-stage-dynamics}.

\section{Limitations and Future Directions}
\label{sec:limitations}

Despite the promising results, reliable route verification remains an open
challenge in multi-step retrosynthesis. In silico checks can serve as proxies
for reaction feasibility, but cannot establish experimental success, yield,
selectivity, reproducibility, or operational safety.

Future work could connect long-horizon LLM agents to automated wet-lab platforms, allowing
proposed routes and reaction conditions to be evaluated against measured outcomes and practical
laboratory constraints. Experimental results could then guide planning and revision over successive
synthesis cycles, allowing long-horizon agents to base future decisions on real experimental results
rather than on in silico assessment alone. Further discussion appears in Appendix~\ref{app:limitations}.

\section{Conclusion}
\label{sec:conclusion}

In this paper, we present Solvent, an 8B LLM that frames multi-step
retrosynthesis as a long-horizon task. Solvent iteratively plans routes, checks
precursor availability, and revises its decisions based on inventory feedback.
A two-round post-training curriculum develops this capability. Across four
benchmarks, Solvent rivals frontier LLMs and consistently outperforms prior
search-based retrosynthesis systems. Case studies further show Solvent exhibiting
strategic chemical reasoning, including protecting-group schemes and convergent
routes where appropriate, highlighting Solvent's chemical insight and the
interpretability of its proposals. Together, Solvent offers a promising long-horizon
paradigm for multi-step retrosynthesis and a path to further scaling.

\clearpage
\section*{AI Use Statement}
In this work, we used generative AI tools for generating synthetic training data, assisting with
coding, and polishing the manuscript. We have not used generative AI tools to develop theoretical
models or conceptual frameworks, formulate mathematical claims, propose or refine hypotheses,
design or provide feedback on research methodology or experiments, clean
or reformat datasets, support qualitative or thematic data analysis, or interpret results; tasks
involving mathematical proofs are not applicable to this work. We have reviewed all AI-assisted
work. We take responsibility for the final content of this work, including text, claims or artifacts
produced with the aid of generative AI.

\section*{Ethics Statement}
This research develops an LLM-based method for multi-step retrosynthesis and does not involve
human subjects or sensitive personal data. All experiments use established chemistry datasets and
benchmarks widely adopted in retrosynthesis research. The proposed method does not involve
demographic attributes or decisions about individuals and therefore raises no direct concerns related
to discrimination or human-subject bias. The authors declare no conflicts of interest or competing
financial interests related to this work. The research follows established practices in machine
learning and computational chemistry.

\section*{Reproducibility Statement}
Appendix~\ref{app:experimental-settings} documents the benchmarks, inventory processing,
evaluation criteria, baseline configurations, subset construction, and planner-budget accounting.
Appendix~\ref{app:training-data} details the training data sources, prompts, filtering procedures, and
target curricula, while Appendix~\ref{app:training-recipe} provides the complete post-training
configurations. Additional ablations and analyses are reported in
Appendix~\ref{app:additional-experiments}, and representative trajectories and their chemical review
are provided in Appendix~\ref{app:trajectories}.
The training and evaluation code is available at
\url{https://github.com/alden199/Solvent}, and the model weights are available at
\url{https://huggingface.co/Alden19910/Solvent}.

\section*{Acknowledgments}
We thank Prof.~Mingyue Zheng, Yuqi Shi, Chuanlong Zeng, and Wei Zhang from the Shanghai Institute of
Materia Medica, Chinese Academy of Sciences, for their assistance with data, as well as for helpful
discussions and valuable feedback.

\bibliography{iclr2026_conference}
\bibliographystyle{iclr2026_conference}

\appendix
\section{Additional Related Work}
\label{app:related-work}

\paragraph{Single-step retrosynthesis models.}
Neural retrosynthesis was first developed around models that predict candidate precursors for an
individual product. Conditional Graph Logic Networks learn when reaction templates are applicable,
whereas LocalRetro combines local reaction templates with global molecular
context~\citep{NEURIPS2019_0d2b2061,Chen_2021}. Other approaches reduce dependence on explicit reaction
templates through representation design or structured generation. Root-aligned SMILES reduces the
sequence discrepancy between products and reactants, while Graph2Edits generates precursors through
autoregressive molecular-graph edits~\citep{Zhong_2022,Zhong_2023}. These models provide the local
reaction proposals on which many multi-step planners operate.

\paragraph{Classical multi-step retrosynthesis planning.}
Multi-step planners traditionally combine a single-step model with an explicit search procedure.
Early neural approaches used Monte Carlo tree search with learned expansion and filtering policies,
while Retro* introduced neural-guided A*-like search over an AND--OR
tree~\citep{segler-etal-2018-mcts,pmlr-v119-chen20k}. Subsequent work improved this paradigm by learning
from successful planning experience, replacing rollouts with experience-guided search, and jointly
optimizing single-step predictions with route-level value
estimates~\citep{pmlr-v139-kim21b,Hong2023,pmlr-v202-liu23as}. DESP further introduced
goal-constrained bidirectional search for routes that must contain specified starting
materials~\citep{NEURIPS2024_cd091a4d}. Despite their different search objectives and heuristics, these methods
construct complete routes by repeatedly selecting and expanding externally proposed single-step
reactions. A complementary direction generates entire routes directly. DirectMultiStep uses a
task-specific encoder--decoder Transformer to generate a complete multi-step route as a structured
sequence~\citep{Shee_2025}. In Solvent, complete-route generation is embedded in an LLM policy that
evaluates and revises its proposals through inventory interaction.

\paragraph{Reasoning, bootstrapping, and tool use.}
Chain-of-thought prompting established intermediate reasoning traces as an effective interface for
complex problem solving~\citep{NEURIPS2022_9d560961}. Fine-tune-CoT transfers such traces from a
larger teacher, while STaR bootstraps reasoning from successful model-generated
rationales~\citep{ho-etal-2023-large,NEURIPS2022_639a9a17}. Toolformer further trains language
models to decide when and how to invoke external APIs~\citep{NEURIPS2023_d842425e}.
Complementary work on test-time computation shows that allocating additional inference effort can
substantially improve reasoning when it is paired with suitable generation and selection
procedures~\citep{snell-etal-2025-scaling}. These foundations motivate models that sustain long-horizon
reasoning while incorporating external evidence into subsequent decisions.

\section{Agentic Reinforcement Learning}
\label{app:agentic-rl}

Agentic reinforcement learning optimizes an LLM over complete multi-turn trajectories, allowing
later decisions to condition on observations from earlier interactions~\citep{pmlr-v235-zhou24t}.

\paragraph{Agent--environment interaction.}
ReAct organizes interactive problem solving as an interleaving of reasoning, actions, and
observations, so that information returned by an environment can inform subsequent
decisions~\citep{yao-etal-2023-react}. For a task input $x$, let $h_t$ denote the interaction history
before turn $t$, and let $r_t$, $a_t$, and $o_t$ denote the reasoning, action, and observation at that
turn. An interaction step is written as
\begin{equation}
    (r_t,a_t)\sim\pi_\theta(\cdot\mid x,h_t),
    \qquad
    o_t=\mathcal{E}_{\mathrm{env}}(a_t),
    \qquad
    h_{t+1}=h_t\oplus(r_t,a_t,o_t),
    \label{eq:agent-environment-trajectory}
\end{equation}
where $\mathcal{E}_{\mathrm{env}}$ maps an action to an observation and $\oplus$ appends the current turn to the
interaction history. For Solvent, $\mathcal{E}_{\mathrm{env}}$ is the building-block inventory, which returns
precursor-availability observations for a submitted route.

\paragraph{Group-relative policy optimization.}
For each input $x\sim\mathcal{D}$, the behavior policy $\pi_{\theta_{\mathrm{old}}}$ samples a group of
$G$ trajectories $\{\tau_i\}_{i=1}^{G}$. Let $L_i$ be the number of policy-generated tokens in
$\tau_i$, $y_{i,t}$ its $t$-th token, and $H_{i,t}$ the interaction history before that token. Let
$R_i$ be the trajectory reward, and let $\bar R$ and $\sigma_R$ be the mean and standard deviation
of rewards within the group. GRPO estimates the advantage from these relative rewards and applies
importance sampling and clipping at the token level~\citep{shao-etal-2024-deepseekmath}:
\begin{equation}
\begin{aligned}
    \mathcal{J}_{\mathrm{GRPO}}(\theta)
    &=\mathbb{E}\!\left[
      \frac{1}{G}\sum_{i=1}^{G}\frac{1}{L_i}\sum_{t=1}^{L_i}
      \min\!\left(
        w_{i,t}(\theta)\widehat{A}_i,
        \operatorname{clip}\!\left(w_{i,t}(\theta),1-\epsilon,1+\epsilon\right)
        \widehat{A}_i
      \right)
    \right],\\
    w_{i,t}(\theta)
    &=\frac{\pi_\theta(y_{i,t}\mid H_{i,t})}
           {\pi_{\theta_{\mathrm{old}}}(y_{i,t}\mid H_{i,t})},
    \qquad
    \widehat{A}_i
    =\frac{R_i-\bar R}{\sigma_R+\varepsilon_A}.
\end{aligned}
\label{eq:grpo-background}
\end{equation}
Here, $\epsilon$ is the token-level clipping bound and $\varepsilon_A=10^{-6}$ prevents division by
zero. Under GRPO, every token in a trajectory shares the same group-relative advantage, while
importance ratios and clipping are computed token by token.

\paragraph{Sequence-level policy optimization.}
GSPO retains group-relative advantage estimation but computes a length-normalized sequence ratio and
applies clipping at the sequence level~\citep{zheng-etal-2025-gspo}. Let $\epsilon_l$ and $\epsilon_h$
denote the lower and upper sequence-level clipping bounds. Its objective is
\begin{equation}
    \mathcal{J}_{\mathrm{GSPO}}(\theta)
    =\mathbb{E}_{x\sim\mathcal{D},\{\tau_i\}_{i=1}^{G}\sim\pi_{\theta_{\mathrm{old}}}(\cdot\mid x)}\!\left[
        \frac{1}{G}\sum_{i=1}^{G}
        \min\!\left(
            s_i(\theta)\widehat{A}_i,
            \operatorname{clip}\!\left(s_i(\theta),1-\epsilon_l,
            1+\epsilon_h\right)\widehat{A}_i
        \right)
      \right].
    \label{eq:gspo-objective}
\end{equation}
The sequence-level ratio is
\begin{equation}
    s_i(\theta)
    =\exp\!\left[
        \frac{1}{L_i}\sum_{t=1}^{L_i}
        \log\frac{\pi_\theta(y_{i,t}\mid H_{i,t})}
                       {\pi_{\theta_{\mathrm{old}}}(y_{i,t}\mid H_{i,t})}
    \right].
    \label{eq:gspo-ratio}
\end{equation}
\section{Experimental Settings}
\label{app:experimental-settings}

\subsection{Benchmarks and Evaluation}
\label{app:benchmarks}

\paragraph{Benchmark composition.}
Our evaluation spans patent-derived synthesis routes and structurally complex bioactive targets.
USPTO-190 contains 190 targets with reference routes extracted from USPTO-Full~\citep{pmlr-v119-chen20k}. ChEMBL-1000 comprises 1,000 compounds sampled from the ChEMBL
bioactivity database and provides no reference routes~\citep{pmlr-v202-liu23as,gaulton2012chembl}. The two Pistachio benchmarks were constructed from
patent routes under constraints on reaction overlap and pairwise target similarity~\citep{NEURIPS2024_cd091a4d}. Pistachio Reachable contains shorter, comparatively tractable routes,
whereas Pistachio Hard emphasizes longer and deeper synthesis plans. Table~\ref{tab:benchmark-details}
summarizes the resulting benchmark statistics. Reference-route statistics characterize the datasets only
and are not used to score predicted routes.

\begin{table}[H]
\centering
\caption{\textbf{Summary of the multi-step retrosynthesis benchmarks.} Reaction counts and depths
refer to the provided reference routes; these statistics are unavailable for ChEMBL-1000.}
\label{tab:benchmark-details}
\footnotesize
\setlength{\tabcolsep}{10pt}
\renewcommand{\arraystretch}{1.08}
\begin{tabular}{lccc}
\toprule
Benchmark & Targets & \shortstack[c]{Avg.\ reactions} & \shortstack[c]{Avg.\ depth} \\
\midrule
USPTO-190 & 190 & 6.7 & 6.0 \\
ChEMBL-1000 & 1,000 & -- & -- \\
Pistachio Reachable & 150 & 5.5 & 5.4 \\
Pistachio Hard & 100 & 7.5 & 7.2 \\
\bottomrule
\end{tabular}
\end{table}

\paragraph{Training--evaluation separation.}
We remove molecular overlaps with the four evaluation benchmarks from every training stage and
further assess structural proximity beyond exact identity. For each test target, we compute its
maximum Tanimoto similarity to any retained training target using radius-2, 2,048-bit Morgan
fingerprints. A target is scaffold-novel when its canonical Bemis--Murcko scaffold is absent from
all retained training targets; targets with empty acyclic scaffolds are excluded.
Table~\ref{tab:train-test-separation}
summarizes the resulting diagnostics.

\begin{table}[H]
\centering
\caption{\textbf{Structural separation between training and evaluation targets.} Similarities are
computed against the union of retained targets from all training stages. Success uses the same fixed
single-trajectory evaluation as Table~\ref{tab:llm-baselines}.}
\label{tab:train-test-separation}
\footnotesize
\setlength{\tabcolsep}{9pt}
\renewcommand{\arraystretch}{1.08}
\begin{tabular}{lc}
\toprule
Diagnostic & Value \\
\midrule
Evaluation targets & 1,440 \\
Exact target overlap & 0 \\
Median maximum Morgan similarity & 0.466 \\
Targets with maximum similarity $\geq 0.8$ & 5.28\% \\
Targets with novel Murcko scaffolds & 81.49\% \\
Success on novel-scaffold targets & 60.26\% \\
\bottomrule
\end{tabular}
\end{table}

The structural separation remains substantial beyond exact deduplication. In particular, only
5.28\% of targets have a close training neighbor at similarity $\geq 0.8$. Moreover, 81.49\% of
scaffold-bearing targets have no Murcko-scaffold match in the training data, and Solvent retains a
60.26\% benchmark-macro success rate on this novel-scaffold subset. Among the 410 targets below
0.4 maximum similarity, its
target-pooled success rate remains 55.37\%. These results support Solvent's generalization beyond
structurally similar training examples, rather than mere memorization of the training data.

\paragraph{Building-block inventory.}
Following prior evaluations of search-based retrosynthesis systems~\citep{NEURIPS2025_66977686,song-etal-2026-aot}, our building-block inventory comprises
commercially available compounds from eMolecules (release dated April 1, 2026). We freeze the inventory before evaluation and
apply the same lookup procedure to all methods evaluated in our pipeline. The prior search-based retrosynthesis systems
likewise follow the eMolecules-based inventory protocol of the corresponding benchmark. Each entry
is reduced to its parent fragment, neutralized where possible, and represented by canonical isomeric
SMILES; predicted route leaves undergo identical normalization before lookup. We use the inventory
to check precursor availability during evaluation and provide feedback during Solvent's multi-turn interaction.

\paragraph{Route-level evaluation.}
Let $d^-_{\mathcal T}(v)$ denote the in-degree of molecule $v$ in the product-to-precursor route
$\mathcal T=(\mathcal V,\mathcal E)$. We first define structural and stock success as
\begin{equation}
    \begin{aligned}
    S_{\mathrm{route}}(\mathcal T;x,\mathcal B)
    ={}&\mathbf{1}\!\left[\operatorname{root}(\mathcal T)=x,\ |\mathcal E|>0\right]\\
       &\times\mathbf{1}\!\left[\mathcal T\ \text{is connected and acyclic}\right]\\
       &\times\prod_{v\in\mathcal V\setminus\{x\}}\mathbf{1}\!\left[d^-_{\mathcal T}(v)=1\right]
        \prod_{\ell\in\mathcal L(\mathcal T)}\mathbf{1}\!\left[\ell\in\mathcal B\right].
    \end{aligned}
    \label{eq:route-evaluation}
\end{equation}
Here, $\mathbf{1}[\cdot]$ denotes the indicator function, equal to one when its condition holds and zero otherwise.
This criterion requires a nontrivial decomposition rooted at the target, a unique parent for every
non-root molecule, and termination entirely in the fixed inventory.

Solvent constructs complete synthesis plans directly rather than confining its reasoning to reaction
candidates enumerated by a single-step model, allowing it to consider a broader range of
transformations. Planning the route as a whole allows the model to coordinate global synthesis
strategy with individual disconnection decisions, thereby better exploiting
its capacity for multi-step reasoning.

\paragraph{Step-level chemical-validity check.}
For each proposed step $e=(p,\mathcal R)$, RDKit first parses and canonicalizes the product $p$ and
the precursor set $\mathcal R$. We assess each step through two complementary checks on reaction patterns
and atom sources. The reaction-support term $c_{\mathrm{sup}}(e)$ checks
whether the product and precursors match the corresponding substructures
of an ASKCOS reaction template~\citep{tu2025askcos}. This substructure-based
check accommodates new substrate combinations.
The atom-source term $c_{\mathrm{atom}}(e)$ checks whether the precursors
contain enough atoms of key elements to account for the product.
Both terms are binary indicators, equal to one when the corresponding check passes and zero otherwise.
Together, these checks allow flexibility in substrate choice while
screening for unsupported reaction patterns and missing atom sources. We define
$c(e)=c_{\mathrm{sup}}(e)c_{\mathrm{atom}}(e)$. The reported success indicator is
\begin{equation}
    S(\mathcal T;x,\mathcal B)
    =S_{\mathrm{route}}(\mathcal T;x,\mathcal B)
     \prod_{e\in\mathcal E} c(e).
    \label{eq:strict-route-evaluation}
\end{equation}
Thus, every step must pass both checks. Together, these criteria provide a practical assessment of
route plausibility, although experimental feasibility can ultimately be established only through
laboratory validation. Developing reliable
computational proxies for experimental outcomes, including yield, selectivity, and competing
reactions, remains an open challenge in synthesis planning~\citep{tu2025askcos,roh2026higher} and
an important direction for future work.

\subsection{Evaluation Configurations and Search Budgets}
\label{app:baseline-budget}

\paragraph{Evaluated LLMs.}
Our LLM comparison includes four proprietary frontier LLMs: Claude Opus
4.8~\citep{claude-opus-4-8-2026}, Gemini 3.5 Flash~\citep{gemini-3-5-flash-2026}, Grok
4.5~\citep{grok-4-5-2026}, and GPT-5.5~\citep{gpt-5-5-2026}. The open-weight frontier models are
GLM-5.2~\citep{glm-5-2-2026}, Kimi K2.5~\citep{kimi-k-2-5-2026},
DeepSeek-V4-Pro~\citep{deepseek-v4-pro-2026}, and Kimi K3~\citep{kimi-k3-2026}; the mid-scale
general-purpose models are Qwen3-8B~\citep{qwen3-technical-report-2025},
Qwen3.5-9B~\citep{qwen3-5-9b-2026}, and Qwen3.6-Flash~\citep{qwen3-6-2026}. We additionally
include $R^3$, an 8B LLM specialized for multi-step route generation~\citep{wang-etal-2026-r3}. All evaluated LLMs receive the same target-only prompt and inventory
interface and are evaluated with one generation per target, up to five reasoning-and-action turns,
and a maximum response length of 32,000 tokens. Temperature, top-$p$, and model-specific
reasoning controls follow each provider's recommended defaults. All outputs are scored by the same
route parser, inventory, and chemical-validity criteria.

\paragraph{Cost-controlled subset.}
To ensure fair and representative comparisons under limited evaluation budgets, all entries marked
with $^\dagger$ in Table~\ref{tab:llm-baselines} are evaluated on the same frozen, model-independent
subset. Structure- and complexity-stratified sampling preserves the chemical diversity of each
benchmark while holding the evaluated targets fixed across models. The resulting manifest contains
160 targets from ChEMBL-1000, 70 from Pistachio Hard, 80 from Pistachio Reachable, and 90 from
USPTO-190.

For each benchmark $D_d=\{x_i\}_{i=1}^{N_d}$, we compute a radius-2, 2048-bit Morgan fingerprint
and six molecular descriptors: heavy-atom, ring, heteroatom, rotatable-bond, and stereocenter counts,
and topological polar surface area. Let $P_{dj}(x_i)\in[0,1]$ denote the within-benchmark percentile
rank of target $x_i$ under descriptor $j$. We define its complexity score as
\begin{equation}
    \kappa_d(x_i)=\frac{1}{6}\sum_{j=1}^{6}P_{dj}(x_i).
    \label{eq:subset-complexity}
\end{equation}
Targets are first partitioned by exact Tanimoto $k$-medoids clustering of their fingerprints, using
eight clusters for ChEMBL-1000 and four for each remaining benchmark. Each cluster is then divided
at its median $\kappa_d$ into lower- and higher-complexity strata. Targets containing bridgehead or
spiro atoms form dedicated lower- and higher-complexity strata so that these uncommon ring
topologies remain represented.

Let $D_{dh}$ be stratum $h$, with population $N_{dh}$, and let $H_d$ be the number of nonempty
strata. Given dataset quota $n_d$, we assign one target to every stratum and distribute the remaining
quota approximately in proportion to residual stratum capacity. Equivalently, constrained
largest-remainder rounding is applied to
\begin{equation}
    q_{dh}=1+(n_d-H_d)\frac{N_{dh}-1}{N_d-H_d},
    \qquad
    \sum_{h=1}^{H_d}n_{dh}=n_d,
    \quad 1\leq n_{dh}\leq N_{dh}.
    \label{eq:subset-allocation}
\end{equation}
We sample $n_{dh}$ targets uniformly without replacement within each stratum using deterministic
seeds derived from the global seed 20260713, dataset, and stratum. The final 400-target manifest is
fixed across all subset-evaluated models. For model $m$, the reported rate on dataset $d$ is the
empirical mean on this shared subset,
\begin{equation}
    \widehat p_{m,d}=\frac{1}{n_d}\sum_{x_i\in\mathcal S_d}
    S(\widehat{\mathcal T}_{m,i};x_i,\mathcal B),
    \qquad
    \widehat p_m=\frac{1}{4}\sum_{d=1}^{4}\widehat p_{m,d},
    \label{eq:subset-success}
\end{equation}
where $\mathcal S_d=\bigcup_h\mathcal S_{dh}$ and $S$ is defined in
Eq.~\ref{eq:strict-route-evaluation}. This design retains broad structural and complexity coverage
within each benchmark while enabling the direct, target-matched comparison reported in
Table~\ref{tab:llm-shared-subset}.

On exactly the same 400 targets, Solvent reaches a 63.26\% macro average, within 0.31 percentage
points of GPT-5.5 and 0.23 points above Kimi K3, while exceeding the same-sized $R^3$ by
43.08 points. The target-matched comparison therefore preserves the conclusion of
Table~\ref{tab:llm-baselines}: Solvent is competitive with frontier LLMs and substantially stronger
than mid-scale alternatives. Its relative standing is not attributable to differences in the
evaluated target sets.

\begin{table}[!htbp]
\centering
\small
\setlength{\tabcolsep}{2.3pt}
\renewcommand{\arraystretch}{1.12}
\renewcommand{\tabularxcolumn}[1]{m{#1}}
\caption{\textbf{LLM performance on the shared 400-target subset.} Every model is evaluated on the
same frozen targets using the success criterion in Eq.~\ref{eq:strict-route-evaluation}. We report
benchmark-specific success rates and their unweighted macro average. Em dashes denote undisclosed
parameter counts. Gray shading marks the best result among frontier LLMs, and boldface marks the
best result among mid-scale LLMs.}
\label{tab:llm-shared-subset}
\begin{tabularx}{\textwidth}{L{3.0cm}C{1.25cm}*{4}{Y}C{2.1cm}}
\toprule
Method &
Params. &
\shortstack[c]{Pistachio\\Hard} &
\shortstack[c]{Pistachio\\Reachable} &
USPTO-190 &
\shortstack[c]{ChEMBL-\\1000} &
Average (\%) \\
\midrule
\multicolumn{7}{l}{\textbf{Frontier LLMs (proprietary)}} \\
\midrule
Claude Opus 4.8 & \textemdash & 38.57 & 77.50 & \frontierbest{54.44} & 55.63 & 56.54 \\
Gemini 3.5 Flash & \textemdash & 42.86 & 72.50 & 52.22 & \frontierbest{65.63} & 58.30 \\
Grok 4.5 & \textemdash & 52.86 & 71.25 & 50.00 & 62.50 & 59.15 \\
GPT-5.5 & \textemdash & \frontierbest{58.57} & \frontierbest{82.50} & \frontierbest{54.44} & 58.75 & \frontierbest{63.57} \\
\midrule
\multicolumn{7}{l}{\textbf{Frontier LLMs (open-weight)}} \\
\midrule
GLM-5.2 & $753\,\mathrm{B}$ & 1.43 & 6.25 & 2.22 & 5.00 & 3.73 \\
Kimi K2.5 & $1\,\mathrm{T}$ & 27.14 & 36.25 & 25.56 & 19.38 & 27.08 \\
DeepSeek-V4-Pro & $1.6\,\mathrm{T}$ & 42.86 & 58.75 & 41.11 & 46.88 & 47.40 \\
Kimi K3 & $2.8\,\mathrm{T}$ & 54.29 & 80.00 & 52.22 & \frontierbest{65.63} & 63.03 \\
\specialrule{\lightrulewidth}{\aboverulesep}{0.6pt}
\specialrule{\lightrulewidth}{0pt}{\belowrulesep}
\multicolumn{7}{l}{\textbf{Mid-scale LLMs}} \\
\midrule
Qwen3-8B & $8\,\mathrm{B}$ & 0.00 & 0.00 & 1.11 & 0.00 & 0.28 \\
Qwen3.5-9B & $9\,\mathrm{B}$ & 0.00 & 1.25 & 0.00 & 0.00 & 0.31 \\
Qwen3.6-Flash & $35\,\mathrm{B}$ & 10.00 & 22.50 & 8.89 & 3.13 & 11.13 \\
$R^3$ & $8\,\mathrm{B}$ & 18.57 & 28.75 & 17.78 & 15.63 & 20.18 \\
\rowcolor{solventblue}
Solvent & $8\,\mathrm{B}$ & \compactbest{64.29} & \compactbest{81.25} & \compactbest{50.00} & \compactbest{57.50} & \compactbest{63.26} \\
\bottomrule
\end{tabularx}
\end{table}

\paragraph{Prior search-based retrosynthesis systems.}
Greedy DFS recursively expands the highest-probability reaction returned by a single-step model~\citep{pmlr-v119-chen20k}.
Retro* performs neural-guided A*-like search over an AND--OR tree, combining accumulated reaction
cost with a learned estimate of the remaining synthesis cost; Retro*-0 removes the learned future-cost
estimate~\citep{pmlr-v119-chen20k}. Retro*+ and Retro*+-0 apply the self-improvement procedure of
\citet{pmlr-v139-kim21b} to the single-step models used by Retro* and Retro*-0, respectively, using
successful routes discovered by the planner to improve subsequent reaction proposals. EG-MCTS
learns an experience-guidance network from successful and failed search trajectories and uses it to
guide Monte Carlo tree search~\citep{Hong2023}. PDVN alternates planning and policy updates while
learning separate value functions for molecular synthesizability and route cost~\citep{pmlr-v202-liu23as}. Retro-R1 instead trains an LLM controller to call a plug-in single-step
model, select among its candidate reactions, and accumulate the resulting decisions in a synthesis
graph~\citep{NEURIPS2025_66977686}.

For the Pistachio comparisons, RootAligned is a single-step sequence model based on root-aligned
SMILES and is paired separately with MCTS and Retro*
search~\citep{Zhong_2022,segler-etal-2018-mcts,pmlr-v119-chen20k}. The LLM (MCTS) and LLM
(Retro*) baselines likewise use an LLM for single-step reaction prediction within the corresponding
search algorithm~\citep{pmlr-v267-wang25ag}. LLM-Syn-Planner generates complete routes and
improves a population of candidates through evaluation, selection, and LLM-based mutation~\citep{pmlr-v267-wang25ag};
AOT* generates complete downstream routes for selected molecules, maps valid proposals into an
AND--OR tree, and reuses intermediate structures through systematic tree search; we report its
DeepSeek-V3 configuration~\citep{song-etal-2026-aot}.

\paragraph{Search budget $N$.}
Search methods that use single-step retrosynthesis models, including Retro*, Retro*+, EG-MCTS, PDVN,
and RETRO-R1, define $N$ as the limit on calls to a single-step retrosynthesis model~\citep{pmlr-v119-chen20k,pmlr-v139-kim21b,Hong2023,pmlr-v202-liu23as,
NEURIPS2025_66977686}. Route-level LLM planners, such as LLM-Syn-Planner and AOT*, define
$N$ by the number of route-generation or tree-expansion operations in their separate search loops~\citep{pmlr-v267-wang25ag,song-etal-2026-aot}. These two definitions are not interchangeable:
the former counts one-step molecule expansions, whereas the latter counts operations that generate
or update route-level search states. Solvent uses a different workflow: it directly generates and
revises complete routes over multiple turns, without calling a single-step model or using a separate
route-search loop.

To compare Solvent with these methods, we need a common concept that covers the different
counting schemes described above. We adopt the search budget $N$ for this purpose. Both types of
counting fit under this concept because each limits the amount of route-search work that can be
performed before a successful route is found. Search methods that use single-step retrosynthesis models count calls to a
single-step model, each of which expands one selected molecule and returns candidate reactions.
Route-level planners count route-level operations, each of which generates or revises a complete
route or expands an existing route tree.

Following the same principle, we define Solvent's search budget strictly and fairly. Solvent's
budget is the total number of reaction steps across all submitted routes. Counting each submitted
reaction step gives a strict account of route-construction work and prevents complete-route
generation from receiving an artificial budget advantage. For a target $x$, let
$\{\widehat{\mathcal T}_q\}_{q=1}^{Q}$ denote the sequence of $Q$ route submissions across
inventory queries and parallel trajectories, with $q$ indexing submissions.
Let $\mathcal E(\widehat{\mathcal T}_q)$ be the set of reaction steps in submission $q$,
and let $C_k$ be the cumulative number of reaction steps through submission $k$.
The success indicator $S(\mathcal T;x,\mathcal B)$ equals one if the route is structurally valid,
rooted at $x$, terminates entirely in inventory $\mathcal B$, and passes both chemical checks
at every step; otherwise, it equals zero (Eq.~\ref{eq:strict-route-evaluation}). We define
\begin{equation}
    C_k=\sum_{q=1}^{k}\left|\mathcal E(\widehat{\mathcal T}_q)\right|,
    \qquad
    \operatorname{Solved}@N
    =\mathbf{1}\!\left[\exists k\in\{1,\ldots,Q\}:\ C_k\leq N
      \ \land\ S(\widehat{\mathcal T}_k;x,\mathcal B)=1\right].
    \label{eq:controlled-budget}
\end{equation}
A target is counted as solved at budget $N$ only if a successful route is found without exceeding
the cumulative reaction-step cost $N$. Other measures of inference cost cannot provide a common
budget for all compared methods. Token counts are not defined for methods without an LLM, and even
when available they cover only language-model output, excluding single-step neural predictions and
tree-search operations. FLOPs are unavailable for proprietary APIs and, when available, measure
model computation rather than route search. Wall-clock time varies with hardware, batching, serving
infrastructure, and implementation. The reaction-step count therefore offers a common measure of
search effort: each counted step represents a chemically interpretable synthesis hypothesis, and the
cumulative count tracks how much reaction space is explored before a viable route is found.
The search budget also has a clear chemical meaning. Before a proposed route can be tested in the
laboratory, each reaction step needs chemical assessment and concrete reaction conditions~\citep{johnson2023counting,zhang2026proceduret5}. The budget therefore approximates the amount of
reaction-level work involved in screening candidate routes, which is relevant to practical synthesis
planning.
Appendix~\ref{app:efficiency}
separately reports the available parameter counts, capped output-token counts, and inventory-query
counts for the evaluated LLMs.

\paragraph{Solvent inference configuration.}
For the final benchmark evaluations, we evaluate Solvent at temperature 0.8 and top-$p$ 1.0,
with up to five interaction turns and a cumulative output limit of 32k tokens.
For Table~\ref{tab:llm-baselines}, we sample one trajectory per target, as for the other LLMs.
For Table~\ref{tab:controlled-budget}, we use parallel sampling to generate a fixed pool of
100 trajectories per target and accumulate their reaction steps, counting a target as solved
if a successful route is found within the search budget $N$.

\section{Post-Training Data Construction}
\label{app:training-data}

\subsection{Building Retrosynthetic Intuition}
\label{app:retrosynthesis-warmup}

This warm-up stage equips the backbone with broad, route-grounded retrosynthetic knowledge before
it is trained to interact with the inventory. We first strictly deduplicate the route corpus used to
train $R^3$~\citep{wang-etal-2026-r3} against all evaluation benchmarks, yielding 104,867
target--route pairs. Each pair is passed to Intern-S1~\citep{bai2025interns1scientificmultimodalfoundation}
through an annotation-generation prompt that includes both the target and its route. Intern-S1 then
produces a target-level structural analysis and a stepwise account of every product-to-precursor
transformation. The resulting annotation is paired with a separate, target-only task prompt to form
the warm-up fine-tuning example. Thus, the route guides annotation generation but is not included in
the input presented to the model being fine-tuned. The complete prompts are provided below; their
placeholders are populated at runtime with the corresponding target and route.

\PromptListing{Annotation-generation prompt}{prompts/intern_s1_warmup_teacher.txt}

\PromptListing{Warm-up training prompt}{prompts/retrosynthesis_student_prompt.txt}

We retain an annotation only when it contains exactly one valid JSON answer block matching the
source route, one ordered step and explanation for every transformation, and a target analysis free
of precursor or disconnection information. We further reject incomplete outputs, conversational or
meta-level language, disclosures of the conditioning route, and any deviations from the source
transformations. Intern-S1's hidden reasoning is removed before dataset assembly; only the visible,
structured annotation is retained. This validation accepts 102,212 annotations for warm-up
fine-tuning.

\begin{table}[H]
\centering
\caption{\textbf{Configuration used to generate retrosynthetic warm-up annotations with Intern-S1.}}
\label{tab:warmup-data-config}
\footnotesize
\setlength{\tabcolsep}{8pt}
\renewcommand{\arraystretch}{1.08}
\begin{tabular}{lc}
\toprule
Setting & Value \\
\midrule
Annotation model & Intern-S1 \\
Source target--route pairs & 104,867 \\
Generations per route & 1 \\
Temperature & 0.7 \\
Top-$p$ / top-$k$ & 1.0 / 50 \\
Reasoning mode & Enabled \\
Valid annotations & 102,212 \\
\bottomrule
\end{tabular}
\end{table}

\subsection{Teacher-Guided Cold Start for Long-Horizon Reasoning}
\label{app:teacher-cold-start}

This stage establishes the model's ability to perform long-horizon reasoning. We assemble 2,200
structurally challenging targets from
USPTO-full~\citep{NEURIPS2019_0d2b2061} after strictly removing all overlaps with the evaluation
benchmarks. Selection uses only molecular structure, balancing molecular size, ring topology,
heteroaromaticity, stereochemistry, flexibility, and functional-group composition while limiting
repeated Murcko scaffolds.
DeepSeek-V4-Pro~\citep{deepseek-v4-pro-2026} then generates eight independent tool-interaction
trajectories per target. Because the model typically produces long reasoning traces, each trajectory
allows at most three inventory-query turns followed by one finalization turn. This retains multiple
opportunities for route revision while keeping the complete trajectory within the 32k-token
generation budget.

The target-only task prompt is identical during rollout generation and fine-tuning. DeepSeek-V4-Pro
generates the reasoning, inventory queries, and final answer, interleaved with observations returned by
the tool; this complete trace becomes the autoregressive target paired with the same user prompt. The
target placeholder in the shared prompt below is populated at runtime, while the inventory tool schema
is supplied through the model's function-calling interface.

\PromptListing{Shared rollout and fine-tuning prompt}{prompts/deepseek_v4_pro_coldstart.txt}

Candidate trajectories undergo three stages of filtering. Route-level checks require a parseable,
nontrivial tree rooted at the target and terminating entirely in available building blocks.
Interaction-level checks require the final route to match a preceding query that received a successful
inventory observation. Finally, each reaction must pass the reaction-support and atom-source checks
defined in Appendix~\ref{app:benchmarks}. We additionally remove malformed, truncated, excessively
long, or repetitive traces.

We rank the surviving candidates using
\begin{equation}
    \begin{aligned}
    Q(\tau)={}&0.22C(\tau)+0.18L(\tau)+0.16D(\tau)+0.18U(\tau)\\
              &+0.18J(\tau)+B_{\mathrm{chem}}(\tau)-0.06R(\tau),
    \end{aligned}
    \label{eq:coldstart-quality}
\end{equation}
where $C$, $L$, and $D$ measure target complexity, normalized trace length, and route depth;
$U$ rewards underrepresented structural attributes; $J$ is the plausibility assessment produced by
DeepSeek-V4-Pro;
$R$ is the repeated-$n$-gram ratio; and $B_{\mathrm{chem}}$ is $0.09$ for the strict chemical tier,
$0.02$ for the relaxed tier, and $0$ otherwise. The strict tier requires every reaction to receive a
high-confidence feasible judgment, whereas the relaxed tier permits a limited fraction of
medium-confidence judgments but no infeasible or low-confidence reactions. For each canonical target,
we first prioritize candidates that pass the LLM-based assessment and then maximize $Q$. The final
set is selected greedily by combining $Q$ with bonuses for underrepresented descriptor groups and
penalties for redundant groups, subject to at most six trajectories per Murcko scaffold. This
procedure yields 700 teacher-generated trajectories for fine-tuning $\pi_{\mathrm{warmed}}$.

\begin{table}[H]
\centering
\caption{\textbf{Configuration used to generate and curate the teacher-guided cold-start trajectories.}}
\label{tab:teacher-cold-start-config}
\footnotesize
\setlength{\tabcolsep}{8pt}
\renewcommand{\arraystretch}{1.08}
\begin{tabular}{lc}
\toprule
Setting & Value \\
\midrule
Teacher model & DeepSeek-V4-Pro \\
Target source & USPTO-full \\
Rollout targets & 2,200 \\
Rollouts per target & 8 \\
Maximum inventory-query turns & 3 \\
Cumulative generation limit & 32k tokens \\
Temperature / top-$p$ & 1.0 / 1.0 \\
Reasoning effort & High \\
Maximum samples per Murcko scaffold & 6 \\
Selected trajectories & 700 \\
\bottomrule
\end{tabular}
\end{table}

\subsection{Target Selection for First-Round RL}
\label{app:first-round-rl-targets}

The first-round RL curriculum comprises 10,000 targets drawn from the USPTO-full
product pool~\citep{NEURIPS2019_0d2b2061} and strictly deduplicated against all evaluation
benchmarks. We retain canonical, single-component products that are absent from the building-block
inventory and exclude structurally trivial molecules. Rather than sampling uniformly, we stratify
targets by structural complexity to span progressively more demanding route-construction problems.
The curriculum increases the representation of polycyclic and fused-ring systems, bridged and spiro
scaffolds, heteroaromatics, stereocenters, flexible linkers, and densely functionalized molecules,
while controlling the concentration of each Murcko scaffold family. This construction preserves
broad scaffold coverage as route complexity increases.

\subsection{Curating Trajectories for Reasoning Bootstrapping}
\label{app:reasoning-bootstrap-data}

\paragraph{Sampling candidate trajectories.}
Reasoning bootstrapping begins with trajectories sampled from $\pi_{\mathrm{seed}}$. We
construct target pools from structurally challenging USPTO-full products~\citep{NEURIPS2019_0d2b2061}
to cover variation in molecular size, ring topology, stereochemistry, flexibility, and functional-group
composition. Before rollout, we canonicalize and deduplicate the targets and remove all exact
overlaps with the evaluation benchmarks. Best-of-$N$ sampling under the configuration in
Table~\ref{tab:reasoning-bootstrap-data} yields 400,000 candidate trajectories. Rollout generation
and SFT use the same target-only task prompt and inventory schema reported in
Appendix~\ref{app:teacher-cold-start}.

\paragraph{Strict trajectory filtering.}
Candidates must satisfy route-level, interaction-level, and reaction-level criteria. The final answer must
define a valid, nontrivial route tree rooted at the target and terminating entirely in available
building blocks. Its interaction trace must be well-formed, agree with the final answer, and include
a successful inventory check of the submitted route. Each reaction must also pass the
chemical-validity assessment in Appendix~\ref{app:benchmarks}. We discard malformed or truncated
outputs, answers emitted before the corresponding inventory observation, and training sequences
that exceed 32k tokens after formatting.

\paragraph{Reasoning-quality selection.}
Following LIMO~\citep{ye-etal-2025-limo}, the reasoning score $q_{\mathrm{reason}}$ combines
elaboration, self-verification, exploratory revision, and adaptive granularity with weights of 0.30,
0.20, 0.25, and 0.25, respectively. The style score $q_{\mathrm{style}}$ combines balanced coverage
of these dimensions, moderate generation uncertainty, tool-use patterns, and successful route
revision after the inventory identifies unavailable precursors, with weights of 0.35, 0.25, 0.20,
and 0.20. We
rank the surviving trajectories by
\begin{equation}
    Q(\tau)=0.85q_{\mathrm{reason}}(\tau)
    +0.15q_{\mathrm{style}}(\tau)-0.30r_{\mathrm{rep}}(\tau),
    \label{eq:bootstrap-ranking}
\end{equation}
where $r_{\mathrm{rep}}$ is the repeated 5-gram ratio. For each canonical target, we retain only the
highest-scoring trajectory. A final decontamination pass against the evaluation benchmarks removes
two additional targets, leaving 9,550 trajectories from 9,550 unique targets.

\begin{table}[H]
\centering
\caption{\textbf{Generation and curation configuration for reasoning bootstrapping.}}
\label{tab:reasoning-bootstrap-data}
\footnotesize
\setlength{\tabcolsep}{8pt}
\renewcommand{\arraystretch}{1.08}
\begin{tabular}{lc}
\toprule
Setting & Value \\
\midrule
Generating policy & $\pi_{\mathrm{seed}}$ \\
Target source & USPTO-full \\
Trajectory sampling & Best-of-$N$ \\
Temperature / top-$p$ & 1.0 / 1.0 \\
Maximum reasoning-and-action turns & 5 \\
Cumulative generation limit & 32k tokens \\
Candidate trajectories & 400,000 \\
Selected trajectories / unique targets & 9,550 / 9,550 \\
\bottomrule
\end{tabular}
\end{table}

\subsection{Target Curriculum for Final-Stage RL}
\label{app:final-rl-targets}

The final-stage RL curriculum concentrates learning on unresolved yet structurally diverse
route-construction problems. We first build a base pool from canonical, single-component
USPTO-full products~\citep{NEURIPS2019_0d2b2061} that are absent from the building-block
inventory. Candidate priority reflects the preceding policy's failure profile, molecular complexity,
and structural attributes, while diversity-aware selection limits repeated Murcko scaffolds. During
base-pool construction, we exclude evaluation targets and targets already selected for SFT.

We then identify 2,909 persistent hard cases whose sampled trajectories during reasoning
bootstrapping yield a 0\% success rate. These targets replace an equal number of
lower-priority examples in the base pool, preserving its overall size and broad scaffold coverage
while shifting the curriculum toward unresolved problems. After final decontamination against the
four evaluation benchmarks, the resulting curriculum contains 10,000 unique targets.

\section{Training Recipe}
\label{app:training-recipe}

We initialize Solvent from the retrosynthesis-pretrained Qwen3-8B checkpoint developed for
$R^3$~\citep{wang-etal-2026-r3} and use it as the backbone for our post-training pipeline.

\subsection{Warm-Up for Retrosynthetic Intuition}
\label{app:warmup-training}

We begin post-training with full-parameter SFT of the retrosynthesis-pretrained 8B backbone on the
curated Intern-S1 annotations described in Appendix~\ref{app:retrosynthesis-warmup}. Of the 102,212
accepted examples, 1,000 are reserved for validation. Table~\ref{tab:warmup-training-config}
summarizes the training configuration. The resulting policy, $\pi_{\mathrm{warmed}}$, initializes the
subsequent teacher-guided cold start.

\begin{table}[H]
\centering
\caption{\textbf{Training configuration for the retrosynthetic-intuition warm-up.}}
\label{tab:warmup-training-config}
\footnotesize
\setlength{\tabcolsep}{6pt}
\renewcommand{\arraystretch}{1.08}
\begin{tabularx}{0.88\linewidth}{@{}X>{\centering\arraybackslash}X@{}}
\toprule
Setting & Value \\
\midrule
Initialization & $R^3$ retrosynthesis-pretrained Qwen3-8B checkpoint \\
Fine-tuning method & Full-parameter SFT \\
Examples / validation split & 102,212 / 1,000 \\
Training epochs & 2 \\
Maximum sequence length & 4,096 tokens \\
Global batch size & 64 \\
Peak learning rate & $3\times10^{-5}$ \\
Learning-rate schedule & Cosine \\
Warm-up ratio & 0.05 \\
Precision & BF16 \\
\bottomrule
\end{tabularx}
\end{table}

\subsection{Teacher-Guided Cold Start}
\label{app:coldstart-training}

We next fine-tune $\pi_{\mathrm{warmed}}$ on the 700 DeepSeek-V4-Pro trajectories curated as
described in Appendix~\ref{app:teacher-cold-start}. Teacher-guided SFT cold-starts the model for
long-horizon reasoning. We select the
checkpoint with the lowest validation loss and use it to initialize first-round RL.
Table~\ref{tab:coldstart-training-config} reports the training configuration.

\begin{table}[H]
\centering
\caption{\textbf{Training configuration for the teacher-guided cold start.}}
\label{tab:coldstart-training-config}
\footnotesize
\setlength{\tabcolsep}{6pt}
\renewcommand{\arraystretch}{1.08}
\begin{tabularx}{0.88\linewidth}{@{}X>{\centering\arraybackslash}X@{}}
\toprule
Setting & Value \\
\midrule
Initialization & $\pi_{\mathrm{warmed}}$ \\
Fine-tuning method & Full-parameter SFT \\
Examples / validation split & 700 / 50 \\
Training epochs & 2 \\
Maximum sequence length & 32,768 tokens \\
Global batch size & 8 \\
Peak learning rate & $2\times10^{-5}$ \\
Learning-rate schedule & Cosine \\
Warm-up ratio & 0.05 \\
Precision & BF16 \\
\bottomrule
\end{tabularx}
\end{table}

\subsection{First-Round RL for Stable Long-Horizon Interaction}
\label{app:first-round-rl-training}

Starting from the teacher-guided cold-start policy, we apply GSPO~\citep{zheng-etal-2025-gspo}
to the 10,000-target curriculum described in Appendix~\ref{app:first-round-rl-targets}. The reward
follows Equation~\ref{eq:seed-reward}, assigning weights of 0.6, 0.2, and 0.2 to route success,
inventory verification, and interaction validity, respectively. Truncated trajectories are excluded
from policy updates, while dynamic group filtering and overlong-response penalties improve training
stability. The resulting policy is denoted by $\pi_{\mathrm{seed}}$.

\begin{table}[H]
\centering
\caption{\textbf{Training configuration for first-round RL.}}
\label{tab:first-round-rl-config}
\footnotesize
\setlength{\tabcolsep}{6pt}
\renewcommand{\arraystretch}{1.08}
\begin{tabularx}{0.88\linewidth}{@{}X>{\centering\arraybackslash}X@{}}
\toprule
Setting & Value \\
\midrule
Initialization & Teacher-guided cold-start policy \\
Optimization algorithm & GSPO \\
Training targets & 10,000 \\
Optimization updates & 50 \\
Rollout targets per batch & 128 \\
Rollouts per target & 8 \\
Global batch size & 512 \\
Maximum interaction turns & 5 \\
Context / response limit & 32,768 / 30,000 tokens \\
Rollout temperature / top-$p$ & 0.9 / 0.9 \\
Learning rate / schedule & $1\times10^{-6}$ / constant \\
Warm-up updates & 20 \\
Precision & BF16 \\
\bottomrule
\end{tabularx}
\end{table}

\subsection{Reasoning Bootstrapping}
\label{app:bootstrap-training}

To construct $\pi_{\mathrm{boot}}$, we return to $\pi_{\mathrm{warmed}}$ and perform full-parameter
SFT on the 9,550 trajectories curated in Appendix~\ref{app:reasoning-bootstrap-data}. We monitor
validation performance and held-out entropy during training, selecting the first-epoch policy to
improve retrosynthesis performance while retaining exploratory capacity. The configuration is
reported in Table~\ref{tab:bootstrap-training-config}.

\begin{table}[H]
\centering
\caption{\textbf{Training configuration for reasoning bootstrapping.}}
\label{tab:bootstrap-training-config}
\footnotesize
\setlength{\tabcolsep}{6pt}
\renewcommand{\arraystretch}{1.08}
\begin{tabularx}{0.88\linewidth}{@{}X>{\centering\arraybackslash}X@{}}
\toprule
Setting & Value \\
\midrule
Initialization & $\pi_{\mathrm{warmed}}$ \\
Fine-tuning method & Full-parameter SFT \\
Examples / validation split & 9,550 / 100 \\
Training epochs & 1 \\
Maximum sequence length & 32,768 tokens \\
Global batch size & 8 \\
Peak learning rate & $2\times10^{-5}$ \\
Learning-rate schedule & Cosine, minimum $1\times10^{-6}$ \\
Warm-up ratio & 0.05 \\
Precision & BF16 \\
\bottomrule
\end{tabularx}
\end{table}

\subsection{Final-Stage RL}
\label{app:final-rl-training}

We optimize $\pi_{\mathrm{boot}}$ with GSPO on the 10,000-target curriculum described in
Appendix~\ref{app:final-rl-targets}, using the gated reward in Equation~\ref{eq:final-reward}.
Dynamic sampling concentrates updates on informative rollout groups, truncated trajectories
are excluded from the policy loss, and overlong responses receive a length penalty. A KL penalty
with coefficient 0.01 further limits policy drift. Table~\ref{tab:final-rl-training-config}
summarizes the core training configuration.

\begin{table}[H]
\centering
\caption{\textbf{Training configuration for final-stage RL.}}
\label{tab:final-rl-training-config}
\footnotesize
\setlength{\tabcolsep}{6pt}
\renewcommand{\arraystretch}{1.08}
\begin{tabularx}{0.88\linewidth}{@{}X>{\centering\arraybackslash}X@{}}
\toprule
Setting & Value \\
\midrule
Initialization & $\pi_{\mathrm{boot}}$ \\
Optimization algorithm & GSPO \\
Training targets & 10,000 \\
Optimization updates & 780 \\
Rollout targets per batch & 64 \\
Rollouts per target & 8 \\
Global batch size & 512 \\
Maximum interaction turns & 5 \\
Context / response limit & 32,768 / 30,000 tokens \\
Rollout temperature / top-$p$ & 1.0 / 1.0 \\
Learning rate / schedule & $1\times10^{-6}$ / constant \\
Warm-up updates & 20 \\
Precision & BF16 \\
\bottomrule
\end{tabularx}
\end{table}

\section{Additional Experiments}
\label{app:additional-experiments}

\subsection{Additional Ablation Studies}
\label{app:additional-ablations}

\subsubsection{Retrosynthetic-Intuition Warm-Up}
\label{app:warmup-ablation}

\textbf{Retrosynthetic-intuition warm-up provides a stronger initialization for long-horizon
reasoning.}
We isolate its contribution by fine-tuning either the retrosynthesis-pretrained
backbone or the warmed-up policy $\pi_{\mathrm{warmed}}$ on the same
reasoning-bootstrapping corpus. Both models use the same training data and optimization settings
and are evaluated at the same training steps. We measure success rate on the four benchmarks,
comprising 1,440 targets, with eight trajectories per target,
temperature 1.0, top-$p$ 1.0, and at most five reasoning-and-action turns.

\begin{figure}[H]
    \centering
    \includegraphics[width=\linewidth]{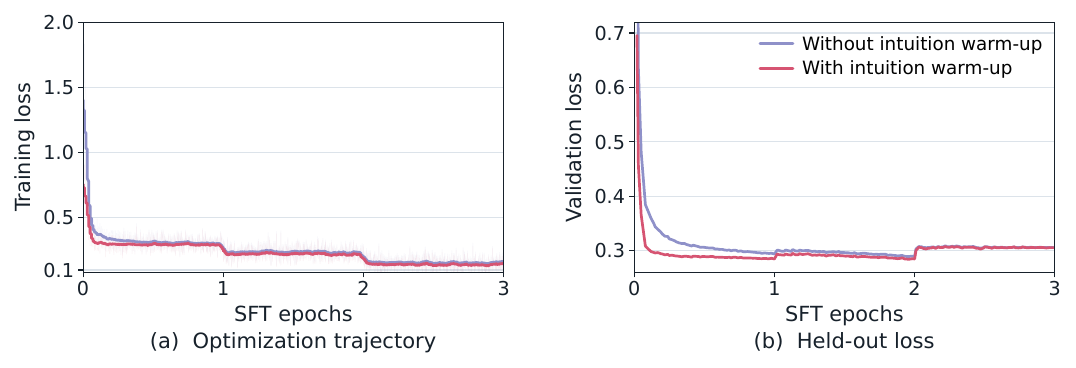}
    \caption{\textbf{Retrosynthetic-intuition warm-up improves the optimization of subsequent
    reasoning-bootstrapping SFT.} Without and with intuition warm-up, training starts from the
    retrosynthesis-pretrained backbone and $\pi_{\mathrm{warmed}}$, respectively.
    (a) Per-step training loss under the two matched initializations;
    pale traces show individual updates and solid curves show centered 61-update averages.
    (b) Validation loss throughout training.}
    \label{fig:warmup-initialization-ablation}
\end{figure}

Figure~\ref{fig:warmup-initialization-ablation} shows that warm-up substantially lowers the initial
training and validation losses and retains a validation advantage through the first two epochs before
the two runs converge. More importantly, this optimization advantage translates into stronger
route-solving performance across training depths (Table~\ref{tab:warmup-initialization-ablation}).

\begin{table}[t]
\centering
\caption{\textbf{Retrosynthetic-intuition warm-up consistently improves success rates under
matched reasoning-bootstrapping SFT.} Results are macro averages across the four benchmarks.}
\label{tab:warmup-initialization-ablation}
\footnotesize
\setlength{\tabcolsep}{5.5pt}
\renewcommand{\arraystretch}{1.08}
\begin{tabular}{c ccc ccc}
\toprule
& \multicolumn{3}{c}{Pass@1 (\%)} & \multicolumn{3}{c}{Pass@8 (\%)} \\
\cmidrule(lr){2-4}\cmidrule(lr){5-7}
SFT epochs & Without warm-up & With warm-up & $\Delta$ & Without warm-up & With warm-up & $\Delta$ \\
\midrule
1 & 24.99 & \textbf{26.42} & +1.43 & 65.24 & \textbf{69.62} & +4.38 \\
2 & 30.21 & \textbf{32.61} & +2.40 & 70.92 & \textbf{73.10} & +2.18 \\
3 & 28.50 & \textbf{32.94} & +4.44 & 74.70 & \textbf{75.26} & +0.56 \\
\bottomrule
\end{tabular}
\end{table}

Warm-up improves both macro Pass@1 and Pass@8 at all three training depths. Together with the lower
early training and validation losses, these consistent gains demonstrate that retrosynthetic-intuition
warm-up provides a more effective initialization for subsequent reasoning-bootstrapping SFT.

\subsubsection{Training Depth for Reasoning Bootstrapping}
\label{app:bootstrap-depth-ablation}

\textbf{Limiting reasoning bootstrapping to one epoch preserves exploration capacity for stronger
downstream RL.}
We isolate the effect of training depth by fine-tuning $\pi_{\mathrm{warmed}}$ on the same curated
trajectory corpus for one, two, or three epochs under an otherwise identical configuration.
Table~\ref{tab:bootstrap-depth-ablation} reports validation loss, entropy,
and task performance after each epoch. We then train the one- and three-epoch policies
with the same final-stage RL configuration to compare their subsequent performance.

\begin{table}[H]
\centering
\caption{\textbf{Validation loss, entropy, and task performance after different numbers of
reasoning-bootstrapping epochs.} Macro success is averaged across the four benchmarks. Bold marks the column maximum,
except for loss, where lower is better. Final parse denotes the rate of parseable final answers.}
\label{tab:bootstrap-depth-ablation}
\footnotesize
\setlength{\tabcolsep}{4.0pt}
\renewcommand{\arraystretch}{1.08}
\begin{tabular}{c c c c c c c}
\toprule
Epoch & Eval loss & \shortstack{SFT\\entropy} & \shortstack{Macro\\P@1 (\%)} &
\shortstack{Macro\\P@8 (\%)} & \shortstack{Rollout\\NLL} & \shortstack{Final parse\\(\%)} \\
\midrule
1 & \textbf{0.2850} & \textbf{0.3217} & 26.42 & 69.62 & \textbf{0.3151} & 71.53 \\
2 & 0.2851 & 0.2830 & 32.61 & 73.10 & 0.2488 & \textbf{81.32} \\
3 & 0.3058 & 0.2586 & \textbf{32.94} & \textbf{75.26} & 0.1927 & 77.01 \\
\bottomrule
\end{tabular}
\end{table}

\begin{figure}[!t]
    \centering
    \includegraphics[width=\linewidth]{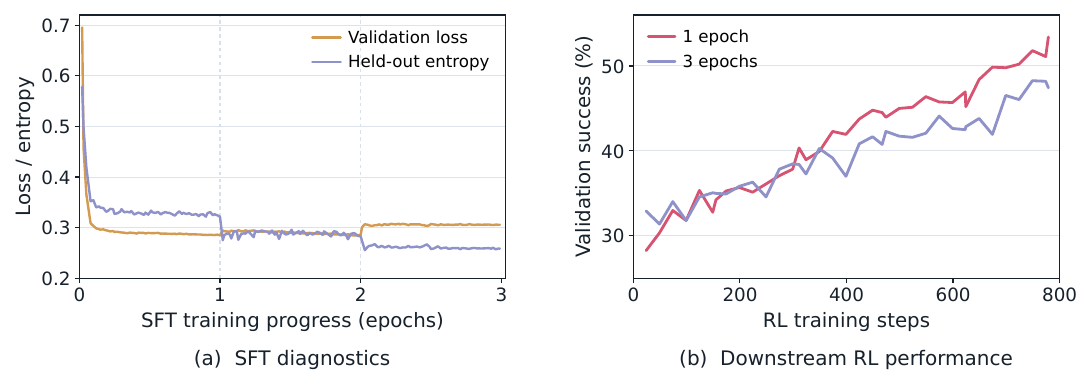}
    \caption{\textbf{One epoch of reasoning bootstrapping yields stronger downstream RL.}
    (a) Validation loss and held-out entropy throughout three epochs of SFT. (b) Validation success
    during matched final-stage RL initialized from the one- and three-epoch policies. RL validation
    uses the default rollout decoding configuration ($T=1.0$, top-$p=1.0$; different from the
    benchmark decoding configuration).}
    \label{fig:bootstrap-depth-ablation}
\end{figure}

Performance before RL improves with training depth: macro Pass@1 rises from 26.42\% to 32.94\%,
while Pass@8 increases from 69.62\% to 75.26\% (Table~\ref{tab:bootstrap-depth-ablation}). However,
the full SFT trace shows that validation loss has largely converged by the end of the first epoch,
whereas held-out entropy continues to decline throughout later training
(Figure~\ref{fig:bootstrap-depth-ablation}(a)). Held-out entropy measures uncertainty in next-token
predictions on fixed validation sequences. Rollout NLL measures the average negative log-probability
of tokens sampled during generation. Both decrease with additional SFT, suggesting reduced
exploration despite higher success rates before RL.

The RL comparison shows that higher success before RL does not necessarily lead to better performance afterward
(Figure~\ref{fig:bootstrap-depth-ablation}(b)). Although the three-epoch policy begins RL with higher
validation success, its improvement plateaus earlier. The one-epoch policy overtakes it as training
proceeds and reaches 53.40\%, compared with 47.43\% from the three-epoch policy.
We therefore select the one-epoch policy for final-stage RL.

\subsubsection{The Large Final-Stage RL Gain Depends on Reasoning Bootstrapping}
\label{app:two-round-curriculum-ablation}

\begin{figure}[!t]
    \centering
    \includegraphics[width=\linewidth]{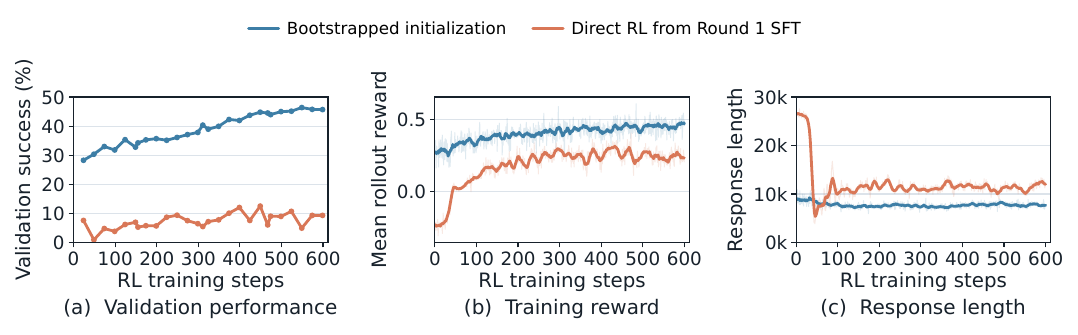}
    \caption{\textbf{Final-stage RL yields large capability gains only after reasoning bootstrapping.}
    (a) Validation success. (b) Mean reward over training rollouts. (c) Mean training-rollout
    response length. Pale traces in (b) and (c) show individual updates, and solid curves show
    centered 11-update averages. Both runs use the default rollout decoding configuration
    ($T=1.0$, top-$p=1.0$); curves are shown through update 600.}
    \label{fig:two-round-curriculum-ablation}
\end{figure}

\textbf{Effective RL benefits from a starting policy that balances task performance with
exploratory capacity~\citep{li-etal-2026-lightweight-sft}.}
We test whether the intermediate training stages establish this starting point by comparing the full
path initialized from $\pi_{\mathrm{boot}}$ with direct application of the same GSPO recipe to the
teacher-guided Round 1 SFT policy. The latter therefore bypasses both Round 1 RL and reasoning
bootstrapping. The target curriculum, reward, optimizer, rollout configuration, and validation
protocol remain fixed. Figure~\ref{fig:two-round-curriculum-ablation} compares the two paths
through 600 training updates.

The results show that the bootstrapped policy provides a stronger starting point for final-stage RL,
reaching 45.69\% validation success compared with 9.31\% when the same RL procedure starts
from the Round 1 SFT policy.

\subsubsection{The Role of Dynamic Sampling in Training Solvent}
\label{app:dynamic-sampling-ablation}

\textbf{Dynamic sampling improves RL training.}
Dynamic sampling in DAPO excludes groups whose sampled outcomes are uniformly correct or
incorrect, concentrating policy updates on prompts that retain a useful learning signal~\citep{yu-etal-2025-dapo}. We adapt this principle to the two route-level outcomes used in our
training procedure. For target $x_i$, let
$\mathcal{G}_i=\{\tau_{i,j}\}_{j=1}^{G}$ denote a group of $G=8$ trajectories. For each trajectory
$\tau_{i,j}$, $r_{i,j}^{\mathrm{stock}}=1$ if its final route is structurally valid and stock-complete,
and $r_{i,j}^{\mathrm{chem}}=1$ if that route is structurally valid and passes the chemical-validity
checks; each indicator is zero otherwise. We define
\begin{equation}
\mathbf{z}_{i,j}
=\left(r_{i,j}^{\mathrm{stock}},r_{i,j}^{\mathrm{chem}}\right)\in\{0,1\}^{2},
\qquad
m_i
=\mathbb{I}\!\left[\exists j:\mathbf{z}_{i,j}\neq(0,0)\right]
 \mathbb{I}\!\left[\exists j:\mathbf{z}_{i,j}\neq(1,1)\right].
\label{eq:dynamic-sampling-filter}
\end{equation}
A group is retained when $m_i=1$. This rule removes groups in which every trajectory fails both
criteria or every trajectory satisfies both, while preserving partial and mixed outcomes. Sampling
continues until each update contains $B=64$ retained groups.

\begin{figure}[!htbp]
    \centering
    \includegraphics[width=\linewidth]{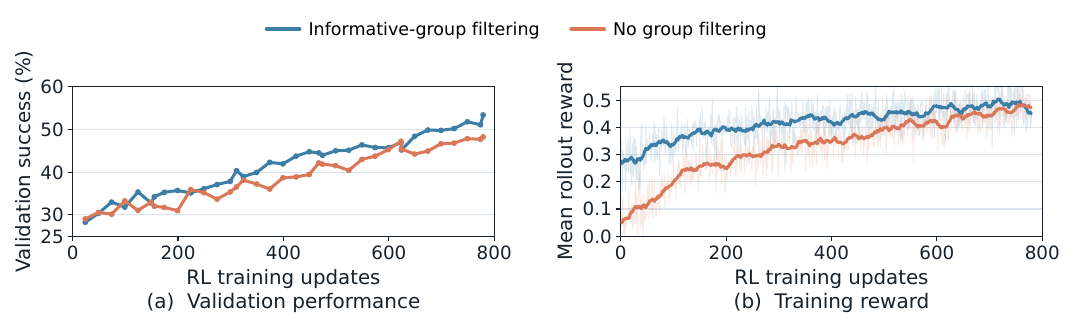}
    \caption{\textbf{Effect of dynamic sampling during final-stage RL.}
    (a) Validation success with and without dynamic sampling. (b) Mean reward over the training
    rollouts retained for each policy update. Pale traces show individual updates, and solid curves
    show centered 21-update averages. Validation uses the default rollout decoding configuration
    ($T=1.0$, top-$p=1.0$).}
    \label{fig:dynamic-sampling-ablation}
\end{figure}

We evaluate the effect of dynamic sampling during final-stage RL by disabling it while keeping
all other training settings fixed. Figure~\ref{fig:dynamic-sampling-ablation} compares their training dynamics.

With dynamic sampling, validation success reaches 53.40\%, compared with 48.26\% without it,
a gain of 5.14 percentage points.

\subsubsection{Adaptive Revision from Availability Feedback}
\label{app:feedback-causal-ablation}

\textbf{Availability feedback drives adaptive revision in long-horizon reasoning.}
To examine the role of availability feedback, we conduct a paired ablation study under three
feedback settings. We select a separate set of 500 targets from USPTO-full, stratified by molecular
size, ring count, and heteroatom count independently of model outputs. After canonicalization, the
set has no exact target overlap with the training data or the four evaluation benchmarks. \emph{True feedback} returns the actual inventory
status of every queried precursor. \emph{Masked feedback} preserves the interaction but withholds
all availability information. \emph{Shuffled feedback} preserves the number of available and
unavailable precursors while permuting their assignments among the queried molecules. The three
conditions use the same checkpoint, decoding configuration, and target-specific sampling seed.

\begin{table}[H]
\centering
\caption{\textbf{Paired intervention on inventory feedback.} Results are computed from one
trajectory for each of 500 held-out targets. The success rate requires a structurally valid,
stock-complete route that passes the chemical-validity assessment.}
\label{tab:feedback-causal-ablation}
\footnotesize
\setlength{\tabcolsep}{5.2pt}
\renewcommand{\arraystretch}{1.08}
\begin{tabularx}{\linewidth}{L{1.45cm}*{5}{>{\centering\arraybackslash}X}}
\toprule
Feedback & \shortstack{Success\\(\%)} & \shortstack{Stock-complete\\(\%)} &
\shortstack{Chemical validity\\(\%)} & \shortstack{Mean response\\tokens} &
\shortstack{Mean inventory\\queries} \\
\midrule
\rowcolor{solventblue}
True & \textbf{68.20} & \textbf{80.40} & \textbf{73.60} & 7,020 & 2.40 \\
Shuffled & 39.00 & 43.80 & 46.60 & 9,158 & 3.27 \\
Masked & 17.60 & 19.20 & 47.20 & 7,891 & 3.63 \\
\bottomrule
\end{tabularx}
\end{table}

As shown in Table~\ref{tab:feedback-causal-ablation}, true feedback improves success by 29.2 percentage points over shuffled feedback
(95\% paired bootstrap CI: 24.4--34.2) and by 50.6 points over masked feedback
(45.8--55.4). These results show that accurate precursor-availability feedback helps
Solvent revise routes successfully.

\FloatBarrier
\subsection{Further Analyses}
\label{app:further-analyses}

\paragraph{Held-out configuration-selection protocol.}
We construct a frozen 500-target validation split from USPTO-full for selecting the inference
configuration. Selection is independent of model outputs and stratified by molecular size, ring
count, and heteroatom count under Murcko-scaffold caps. After canonicalization, we remove targets
present in the building-block inventory and enforce zero exact target overlap with all training
corpora and the four evaluation benchmarks. This split is used only for configuration selection.

\subsubsection{Selecting the Number of Interaction Turns on Held-Out Data}
\label{app:interaction-turn-selection}

\textbf{Held-out selection supports the five-turn inference budget.}
The main evaluation protocol prespecifies a maximum of five interaction turns. We select among
turn limits of 2--5 on the held-out split using one trajectory per target, temperature 0.8, and top-$p$ 1.0;
six- and seven-turn limits provide sensitivity analyses beyond this protocol. Table~\ref{tab:interaction-turn-performance}
reports performance and inference behavior across turn limits.

\vspace*{0pt plus 1fill}
\begin{table}[H]
\centering
\caption{\textbf{Held-out performance and inference behavior across turn limits.}
Blue shading marks the selected configuration, and bold marks the maximum in each performance
column. Tool calls are averaged over all trajectories or successful trajectories only.}
\label{tab:interaction-turn-performance}
\footnotesize
\setlength{\tabcolsep}{3.5pt}
\renewcommand{\arraystretch}{1.08}
\begin{tabular*}{\linewidth}{@{\extracolsep{\fill}}cccccccc@{}}
\toprule
 & \multicolumn{3}{c}{Performance} & \multicolumn{4}{c}{Inference behavior} \\
\cmidrule(lr){2-4}\cmidrule(lr){5-8}
\shortstack{Turn\\limit} & \shortstack{Success\\rate (\%)} &
\shortstack{Stock-complete\\routes (\%)} & \shortstack{Chemically valid\\routes (\%)} &
\shortstack{Mean tool\\calls (all)} & \shortstack{Mean tool calls\\(successful)} &
\shortstack{Mean response\\tokens} & \shortstack{Truncated\\(\%)} \\
\midrule
2 & 27.20 & 28.40 & 27.80 & 1.71 & 1.00 & 4,524 & 0.00 \\
3 & 50.20 & 56.00 & 51.60 & 2.07 & 1.38 & 5,743 & 0.20 \\
4 & 62.80 & 70.40 & 65.40 & 2.33 & 1.71 & 6,563 & 0.00 \\
\rowcolor{solventblue}
5 & 68.00 & 77.40 & 71.80 & 2.48 & 1.92 & 7,063 & 0.20 \\
6 & 69.20 & \textbf{83.60} & 74.00 & 2.49 & 2.10 & 7,294 & 0.20 \\
7 & \textbf{70.60} & 83.40 & \textbf{76.20} & 2.54 & 2.12 & 7,459 & 0.20 \\
\bottomrule
\end{tabular*}
\end{table}

Within the predefined selection range, success increases monotonically from 27.20\% with two turns to
68.00\% with five. Allowing six or seven turns yields smaller gains of 1.2 and 2.6
percentage points, respectively; both paired 95\% bootstrap confidence intervals include zero
($-3.4$--5.6 and $-2.0$--7.2).

Mean inventory use and response length grow gradually as the turn limit increases, while truncation remains at or
below 0.20\% throughout. The small behavioral changes beyond five turns mirror the saturation in
success. A five-turn limit is therefore selected within the prespecified range and fixed before benchmark
evaluation; allowing additional turns yields modest gains.

\subsubsection{Selecting the Sampling Temperature on Held-Out Data}
\label{app:inference-setup}

\textbf{Temperature 0.8 provides the strongest held-out performance with stable generation.}
We vary the temperature from 0.6 to 1.0 with top-$p$ fixed at 1.0, a five-turn limit, and a 32k-token
generation limit. Each setting produces one trajectory per held-out target.
Table~\ref{tab:temperature-analysis} reports the success rate and its two components,
together with generation diagnostics. Sampled-token NLL is the mean negative log-probability of
generated tokens and measures sampled-token surprisal rather than full-vocabulary entropy.

\begin{table}[H]
\centering
\caption{\textbf{Inference performance and generation behavior across sampling temperatures.}
Results are computed on the frozen 500-target validation split. Blue shading marks the selected
configuration; bold marks the highest task metrics and the lowest truncation rate.}
\label{tab:temperature-analysis}
\footnotesize
\setlength{\tabcolsep}{4.0pt}
\renewcommand{\arraystretch}{1.10}
\renewcommand{\tabularxcolumn}[1]{m{#1}}
\begin{tabularx}{\textwidth}{C{0.42cm}*{7}{>{\centering\arraybackslash}X}}
\toprule
& \multicolumn{3}{c}{Performance (\%)} & \multicolumn{4}{c}{Generation behavior} \\
\cmidrule(lr){2-4}\cmidrule(lr){5-8}
$T$ & \shortstack{Success\\rate} & \shortstack{Stock-\\complete} &
\shortstack{Chemical\\validity} & \shortstack{Token\\NLL} &
\shortstack{Response\\tokens} & \shortstack{Inventory\\queries} &
\shortstack{Truncated\\(\%)} \\
\midrule
0.6 & 65.80 & 76.00 & 69.60 & 0.1084 & 7,055 & 2.52 & 0.80 \\
0.7 & 66.80 & 77.20 & 71.40 & 0.1322 & 6,911 & 2.48 & 0.40 \\
\rowcolor{solventblue}
0.8 & \textbf{68.00} & \textbf{77.40} & \textbf{71.80} & 0.1638 & 7,063 & 2.48 & \textbf{0.20} \\
0.9 & 65.00 & 76.80 & 70.80 & 0.2103 & 7,595 & 2.47 & 0.60 \\
1.0 & 63.00 & 73.60 & 70.80 & 0.2847 & 7,958 & 2.43 & \textbf{0.20} \\
\bottomrule
\end{tabularx}
\end{table}

Temperature 0.8 yields the highest success rate, stock-completion rate, and rate of
chemically valid routes. Its 0.20\% truncation rate is tied for the lowest among the evaluated
temperatures, while its mean response length remains close to the lower-temperature settings and
below those at 0.9 and 1.0. Relative to temperature 1.0, it improves paired success by 5.0
percentage points (95\% bootstrap CI: 0.2--9.8). Its point estimate also exceeds those at
temperatures 0.6, 0.7, and 0.9, although the corresponding paired intervals include zero. We
therefore select temperature 0.8 with top-$p$ 1.0 on the held-out split and fix this configuration
for the main results in Tables~\ref{tab:llm-baselines} and~\ref{tab:controlled-budget}.

\subsubsection{Inference Cost}
\label{app:efficiency}

Table~\ref{tab:inference-efficiency} characterizes inference cost using parameter count, mean
output tokens, and mean inventory queries per trajectory. Evaluations marked with $^\dagger$ use
the shared 400-target subset, whereas unmarked results use the full benchmark test sets. Output-token
counts are capped at the shared 32k trajectory budget before averaging.

Solvent achieves comparable average success to Kimi K3 at a smaller model size
and with fewer output tokens: 61.86\% versus 63.03\%, with 8B versus 2.8T
parameters and 9.20k versus 18.45k output tokens per trajectory.

\begin{table}[!t]
\centering
\caption{\textbf{Inference cost of the evaluated LLMs.} Success and usage are macro-averaged;
daggers mark the shared 400-target subset, and em dashes denote undisclosed parameter counts.}
\label{tab:inference-efficiency}
\footnotesize
\setlength{\tabcolsep}{2.0pt}
\renewcommand{\arraystretch}{1.08}
\begin{tabular}{L{4.10cm} C{1.30cm} C{1.50cm} C{1.50cm} C{1.70cm}}
\toprule
Method & Params. & \shortstack{Success\\(\%)} & \shortstack{Output\\tokens} &
\shortstack{Inventory\\queries} \\
\midrule
\multicolumn{5}{l}{\textbf{Frontier LLMs (proprietary)}} \\
\midrule
Claude Opus 4.8$^\dagger$ & \textemdash & 56.54 & 3.57k & 2.73 \\
Gemini 3.5 Flash$^\dagger$ & \textemdash & 58.30 & 20.33k & 2.70 \\
Grok 4.5$^\dagger$ & \textemdash & 59.15 & 14.30k & 1.72 \\
GPT-5.5$^\dagger$ & \textemdash & 63.57 & 5.32k & 2.57 \\
\midrule
\multicolumn{5}{l}{\textbf{Frontier LLMs (open-weight)}} \\
\midrule
GLM-5.2$^\dagger$ & $753\,\mathrm{B}$ & 3.73 & 30.50k & 0.09 \\
Kimi K2.5$^\dagger$ & $1\,\mathrm{T}$ & 27.08 & 9.77k & 2.59 \\
DeepSeek-V4-Pro & $1.6\,\mathrm{T}$ & 42.70 & 23.94k & 2.25 \\
Kimi K3$^\dagger$ & $2.8\,\mathrm{T}$ & 63.03 & 18.45k & 1.61 \\
\midrule
\multicolumn{5}{l}{\textbf{Mid-scale LLMs}} \\
\midrule
Qwen3-8B & $8\,\mathrm{B}$ & 0.16 & 3.49k & 2.27 \\
Qwen3.5-9B & $9\,\mathrm{B}$ & 0.17 & 5.67k & 2.11 \\
Qwen3.6-Flash$^\dagger$ & $35\,\mathrm{B}$ & 11.13 & 29.80k & 0.53 \\
$R^3$ & $8\,\mathrm{B}$ & 19.03 & 2.33k & 0.00 \\
\rowcolor{solventblue}
Solvent & $8\,\mathrm{B}$ & \textbf{61.86} & 9.20k & 2.89 \\
\bottomrule
\end{tabular}
\end{table}

\subsubsection{Complementary In Silico Assessments of Route Quality}
\label{app:orthogonal-route-assessments}

We complement the primary success criterion with three orthogonal post hoc assessments on the
frozen 400-target subset shared by all LLMs in Table~\ref{tab:llm-baselines}. These diagnostics
examine distinct properties of stock-complete routes: structural simplification, support from
documented reaction contexts, and consistency with a learned forward-reaction model. None is used
for training, model selection, or the primary chemical-validity assessment. The ChemCensor and
round-trip diagnostics apply an all-step route criterion. For a route
$\tau=(e_1,\ldots,e_L)$, acceptance is the conjunction
$C(\tau)=\prod_{j=1}^{L}c(e_j)$: a single unsupported transformation rejects the complete route.
Full-route acceptance therefore becomes more stringent with route length. Under an illustrative
independence assumption, a 70\% per-step acceptance rate corresponds to only 16.8\% over five
steps. A low route-level rate therefore does not imply that most individual reactions are
unsupported. At the reaction level, recent out-of-distribution retrosynthesis evaluations report
round-trip rates ranging from 5\% to 30\% and show through manual review that recovery depends on
the coverage of the forward model~\citep{choe2026extrapolation}. For transformations outside the
recorded reaction space, absence of support should therefore be interpreted as insufficient
computational evidence rather than evidence of experimental
failure~\citep{bradshaw2025challenging,choe2026extrapolation}. Taken together, these \textit{in silico}
diagnostics provide complementary evidence of route plausibility, but do not establish experimental
feasibility or substitute for wet-lab validation. Further discussion appears in
Appendix~\ref{app:limitations}.

\paragraph{Synthetic accessibility.}
The SA score estimates molecular accessibility from fragment contributions and structural
complexity, with lower values indicating more accessible structures~\citep{Ertl_2009}. We measure
the decrease from the target score to the mean score of the terminal precursors and the fraction of
retrosynthetic steps whose precursors are, on average, more accessible than their product. Solvent
achieves a target-to-leaf decrease of 0.827, and 84.69\% of its steps reduce the SA score. The
corresponding values are 0.662 and 66.32\% for Kimi K3, and 0.567 and 64.09\% for $R^3$, indicating
that Solvent more consistently decomposes targets into structurally accessible precursors.

\paragraph{ChemCensor.}
ChemCensor evaluates whether a proposed reaction center and its functional-group context are
supported by documented reaction precedents~\citep{zagribelnyy2026singleanswerenoughrethinking}.
We use ChemCensor v1.2.0 and report the number of stock-complete routes for which every step
passes the reaction-center (RC) or functional-group (FG) criterion per 100 evaluated reaction steps.
This normalization rewards complete, consistently supported routes while accounting for the
amount of explicit reaction construction. Solvent obtains 8.62 RC-supported and 4.11 FG-supported
routes per 100 steps, comparable to Kimi K3 (8.03 and 3.65), Claude Opus 4.8 (8.35 and 3.49), and
Grok 4.5 (8.48 and 4.48). Disabling ChemCensor's exact-reaction lookup leaves Solvent's results
unchanged and alters every model's route-level rate by at most 0.31 percentage points. The evidence
therefore reflects compatibility with broader reaction contexts rather than exact database overlap.

\paragraph{Forward round-trip consistency.}
We additionally apply the USPTO-50K-finetuned Chemformer forward model~\citep{Irwin_2022} to
each proposed step and test whether the original product is recovered among its top-20 predictions.
We assess route-level consistency by requiring recovery at every step. Solvent reaches 9.02\%,
exceeding DeepSeek-V4-Pro (7.52\%) and $R^3$ (4.03\%), approaching GPT-5.5 (10.90\%), and
falling within the range of the frontier LLMs in Table~\ref{tab:orthogonal-route-quality}.

\begin{table}[!t]
\centering
\caption{\textbf{Complementary \textit{in silico} assessments on the shared 400-target subset.}
Stock routes are stock-complete. $\Delta$SA is the decrease from target SA to mean leaf SA;
reducing steps have a lower mean precursor SA than product SA. ChemCensor columns report
all-step-supported stock-complete routes per 100 evaluated reaction steps, macro-averaged across
benchmarks. Route RT is the percentage of all 400 targets whose routes pass exact top-20
forward recovery at every step.}
\label{tab:orthogonal-route-quality}
\footnotesize
\setlength{\tabcolsep}{2.0pt}
\renewcommand{\arraystretch}{1.06}
\begin{tabularx}{\linewidth}{L{2.55cm} C{0.95cm}*{6}{>{\centering\arraybackslash}X}}
\toprule
& & & \multicolumn{2}{c}{Synthetic accessibility} &
\multicolumn{2}{c}{ChemCensor} & \multicolumn{1}{c}{Round trip} \\
\cmidrule(lr){4-5}\cmidrule(lr){6-7}\cmidrule(lr){8-8}
Method & Params. & \shortstack{Stock\\routes\\({/}400)} & \shortstack{Target--leaf\\$\Delta$SA} &
\shortstack{Reducing\\steps (\%)} & \shortstack{RC routes\\{/}100 steps} &
\shortstack{FG routes\\{/}100 steps} & \shortstack{Route\\RT@20 (\%)} \\
\midrule
\multicolumn{8}{l}{\textbf{Frontier LLMs (proprietary)}} \\
\midrule
Claude Opus 4.8 & \textemdash & 304 & 0.611 & 70.10 & 8.35 & 3.49 & 14.42 \\
Gemini 3.5 Flash & \textemdash & 300 & 0.607 & 68.60 & 9.89 & 6.02 & 15.75 \\
Grok 4.5 & \textemdash & 295 & 0.618 & 63.89 & 8.48 & 4.48 & 13.06 \\
GPT-5.5 & \textemdash & 289 & 0.539 & 66.57 & 9.98 & 5.33 & 10.90 \\
\midrule
\multicolumn{8}{l}{\textbf{Frontier LLMs (open-weight)}} \\
\midrule
GLM-5.2 & $753\,\mathrm{B}$ & 18 & 0.890 & 73.01 & 22.90 & 12.84 & 1.22 \\
Kimi K2.5 & $1\,\mathrm{T}$ & 235 & 0.539 & 68.23 & 2.42 & 0.87 & 2.50 \\
DeepSeek-V4-Pro & $1.6\,\mathrm{T}$ & 307 & 0.711 & 68.67 & 4.95 & 1.86 & 7.52 \\
Kimi K3 & $2.8\,\mathrm{T}$ & 309 & 0.662 & 66.32 & 8.03 & 3.65 & 13.44 \\
\midrule
\multicolumn{8}{l}{\textbf{Mid-scale LLMs}} \\
\midrule
Qwen3-8B & $8\,\mathrm{B}$ & 13 & 0.787 & 78.24 & 0.00 & 0.00 & 0.00 \\
Qwen3.5-9B & $9\,\mathrm{B}$ & 2 & 0.576 & 75.00 & 0.00 & 0.00 & 0.00 \\
Qwen3.6-Flash & $35\,\mathrm{B}$ & 62 & 0.750 & 78.73 & 6.98 & 1.54 & 0.78 \\
$R^3$ & $8\,\mathrm{B}$ & 100 & 0.567 & 64.09 & 6.53 & 4.75 & 4.03 \\
\rowcolor{solventblue}
\textbf{Solvent} & $8\,\mathrm{B}$ & 293 & 0.827 & 84.69 & 8.62 & 4.11 & 9.02 \\
\bottomrule
\end{tabularx}
\end{table}

\paragraph{Consistency of the primary chemical-validity assessment.}
We assess agreement on the 293 stock-complete Solvent routes, thereby holding inventory completion
fixed. Routes accepted by the primary chemical-validity assessment receive consistently stronger
ChemCensor support and forward-model recovery than routes not accepted by it
(Table~\ref{tab:orthogonal-validator-agreement}). The differences remain positive after joint
stratification by benchmark and route length. These convergent results support the practical
discriminative value of the primary assessment while preserving the distinct scope of each
\textit{in silico} diagnostic.

\begin{table}[!t]
\centering
\caption{\textbf{Orthogonal assessments grouped by primary chemical validity.}
We split 293 stock-complete Solvent routes into routes that pass and routes that do not pass the
primary chemical-validity assessment. Each row reports the mean score of an independent diagnostic
in the two groups. Difference is the mean for primary-pass routes minus the mean for
primary-non-pass routes; 95\% confidence intervals use 10,000 benchmark-stratified bootstrap
resamples.}
\label{tab:orthogonal-validator-agreement}
\footnotesize
\setlength{\tabcolsep}{3.5pt}
\renewcommand{\arraystretch}{1.08}
\begin{tabularx}{\linewidth}{L{4.70cm} C{1.20cm} C{1.20cm} >{\centering\arraybackslash}X}
\toprule
Orthogonal diagnostic & \shortstack{Primary\\pass (\%)} &
\shortstack{Primary\\non-pass\\(\%)} & Difference (95\% CI) \\
\midrule
ChemCensor RC step support & 59.72 & 45.92 & $+13.80$ ($6.65$, $21.15$) \\
ChemCensor FG step support & 45.14 & 29.90 & $+15.24$ ($7.67$, $22.76$) \\
Exact route RT@20 & 14.98 & 0.00 & $+14.98$ ($10.67$, $19.50$) \\
\bottomrule
\end{tabularx}
\end{table}

\subsubsection{Behavioral Dynamics of the Two RL Stages}
\label{app:rl-stage-dynamics}

\begin{figure}[!t]
    \centering
    \includegraphics[width=\linewidth]{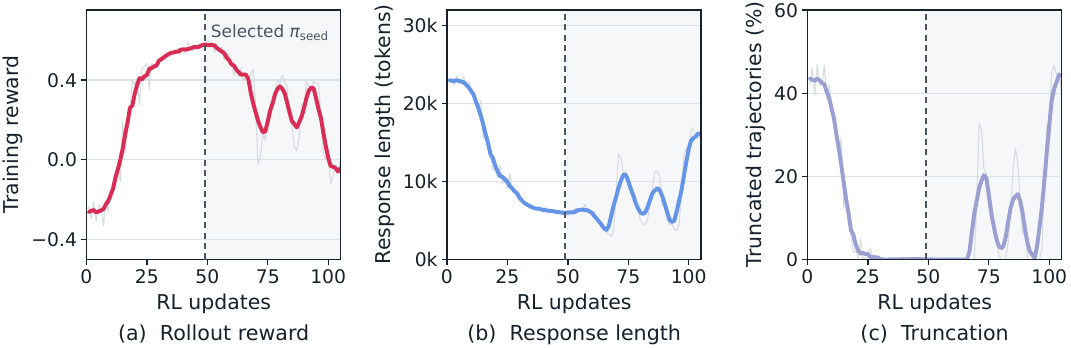}
    \caption{\textbf{Round 1 RL stabilizes long-horizon interaction.} (a) Training reward,
    (b) mean response length, and (c) the fraction of truncated trajectories. Faint lines show
    per-update measurements and colored lines show centered seven-update moving averages. The
    dashed line marks the selected $\pi_{\mathrm{seed}}$; the shaded region denotes the exploratory
    continuation beyond this checkpoint.}
    \label{fig:round1-rl-behavior}
\end{figure}

\textbf{Round 1 RL stabilizes long-horizon interaction.}
Figure~\ref{fig:round1-rl-behavior} traces first-round RL through the checkpoint selected as
$\pi_{\mathrm{seed}}$ and an exploratory continuation beyond it. Across the first 49 updates,
five-update averages show reward rising from $-0.255$ to $0.569$, while mean response length falls
from 22.9k to 6.1k tokens and truncation from 43.2\% to 0.12\%. Continued optimization reverses
these trends. We therefore use the update-49 checkpoint as $\pi_{\mathrm{seed}}$ for subsequent
trajectory collection.

\begin{figure}[!t]
    \centering
    \includegraphics[width=0.88\linewidth]{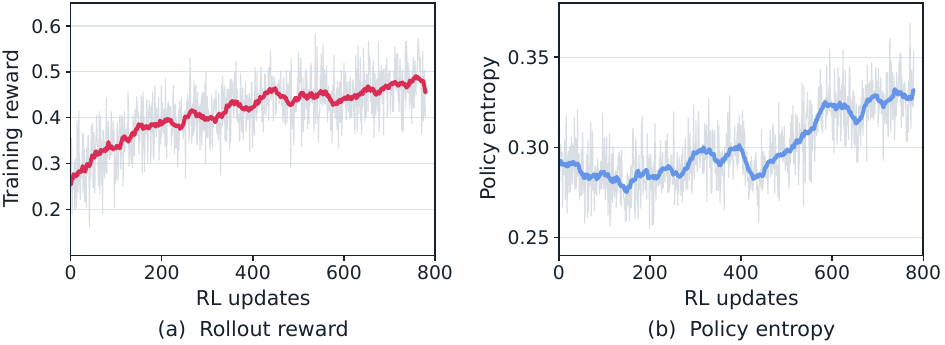}
    \caption{\textbf{Reward and entropy during final-stage RL.} (a) Mean rollout reward and
    (b) policy entropy. Faint lines show per-update measurements and colored lines show centered
    25-update moving averages.}
    \label{fig:final-rl-reward-entropy}
\end{figure}

\textbf{Final-stage RL improves reward while maintaining policy entropy.}
As shown in Figure~\ref{fig:final-rl-reward-entropy}, mean rollout reward increases throughout
final-stage RL, from 0.274 over the first 25 updates to 0.484 over the last 25. Meanwhile, policy
entropy is maintained throughout training as rollout reward improves.

\section{Representative Case Studies}
\label{app:trajectories}

\subsection{Strategic Retrosynthetic Reasoning and Feedback-Guided Revision}
\label{app:case-feedback-revision}

\subsubsection{Inventory-Guided Convergent Route Planning}
\label{app:case-convergent-route}

\begin{figure}[H]
    \centering
    \includegraphics[width=0.52\linewidth]{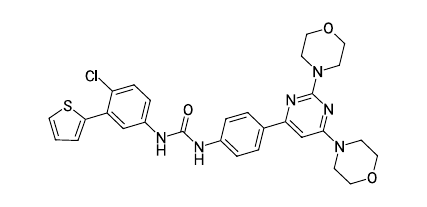}
    \caption{Target molecule for the convergent-route case study.}
    \label{fig:case-convergent-target}
\end{figure}

This example illustrates how Solvent revises a high-level disconnection in response to precursor
availability and then develops both branches of a convergent synthesis. The target is an
unsymmetrical diaryl urea containing bis(morpholinyl)pyrimidine and chlorothiophene fragments
(Figure~\ref{fig:case-convergent-target}).

\noindent\textbf{Target SMILES.}
\begin{lstlisting}[style=tooltrace]
O=C(Nc1ccc(-c2cc(N3CCOCC3)nc(N3CCOCC3)n2)cc1)Nc1ccc(Cl)c(-c2cccs2)c1
\end{lstlisting}

\noindent\textbf{Model trajectory.}
The selected verbatim excerpts retain the structural analysis, inventory-guided revisions, and
final route review; bracketed ellipses mark omitted intermediate analysis and tool payloads.
\ReasoningListing{Selected reasoning trace}{case_studies/chembl_615_trace_excerpt.txt}

The three inventory queries progressively expand the route until every terminal precursor is
available:
\begin{table}[H]
\centering
\caption{Inventory feedback during route revision.}
\label{tab:case-convergent-inventory}
\small
\setlength{\tabcolsep}{7pt}
\renewcommand{\arraystretch}{1.08}
\begin{tabularx}{\linewidth}{L{2.1cm}L{7.7cm}Y}
\toprule
Query & Proposed route boundary & Available leaves \\
\midrule
1 & Aniline and aryl isocyanate & 0/2 \\
2 & Two anilines and CDI & 1/3 \\
3 & Both aniline branches expanded to simpler precursors & 6/6 \\
\bottomrule
\end{tabularx}
\end{table}

\begin{figure}[!t]
    \centering
    \includegraphics[width=\linewidth]{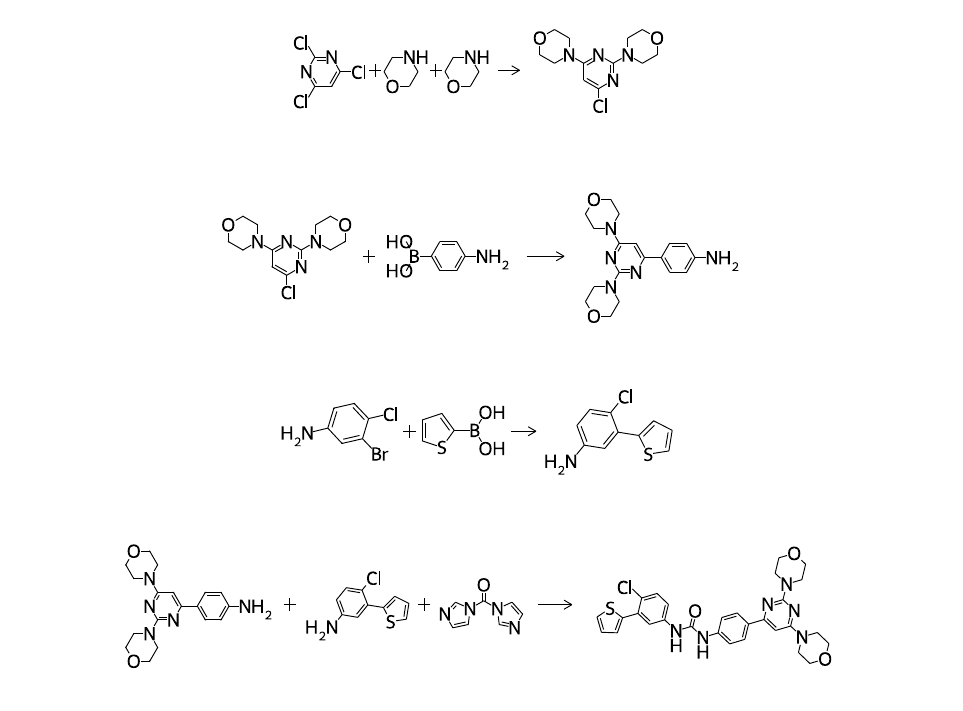}
    \caption{Inventory-verified convergent route rendered in forward synthetic order.}
    \label{fig:case-convergent-route}
\end{figure}

Figure~\ref{fig:case-convergent-route} renders the accepted route in forward synthetic order. The
two aniline fragments are prepared independently and joined through the central urea in the final
step.

\Needspace{8\baselineskip}
\ToolListing{Final answer}{case_studies/chembl_615_final_answer.json}
\noindent\textbf{Manual chemical review.}
The route was also examined transformation by transformation. (1) The electron-deficient
trichloropyrimidine provides a credible substrate for sequential substitution by morpholine, although
the two substitutions would ordinarily require staged condition control. (2) Coupling the residual
chloropyrimidine with 4-aminophenylboronic acid forms the first aryl fragment through a Suzuki
reaction. (3) The second Suzuki reaction exploits the greater reactivity of the aryl bromide while
retaining the aryl chloride in the product. (4) CDI provides a plausible carbonyl source for staged
coupling of the two anilines into the unsymmetrical urea. As
complementary computational evidence, all four transformations pass ChemCensor's reaction-center
criterion, three pass its functional-group criterion, and an
independently trained Chemformer model recovers every product within its top-5 forward predictions.

\noindent\textbf{Interpretation.}
The trajectory begins with a concise urea disconnection, abandons an unavailable isocyanate route,
and independently expands both aryl branches before returning a stock-complete convergent plan.
The value of the example lies in this coordination of global assembly, branch-specific reaction
selection, and revision from inventory evidence rather than in any single isolated transformation.

\FloatBarrier
\subsubsection{Boc Protection Enables Chemoselective Piperazine Functionalization}
\label{app:case-boc-protection}

This example illustrates how inventory feedback prompts a concrete protecting-group strategy.
Solvent initially proposes 1-(4-methoxyphenyl)piperazine as a terminal precursor, but the inventory
reports that it is unavailable. The model then constructs this intermediate from N-Boc-piperazine
and 4-bromoanisole, retaining the Boc group during selective N-arylation and removing it before the
remaining piperazine nitrogen is coupled to the activated maleimide.

\begin{figure}[H]
    \centering
    \includegraphics[width=0.46\linewidth]{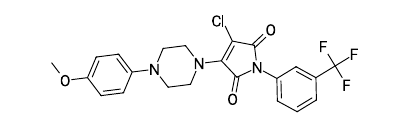}
    \caption{Target molecule for the Boc-protection case study.}
    \label{fig:case-boc-target}
\end{figure}

\noindent\textbf{Target SMILES.}
\begin{lstlisting}[style=tooltrace]
COc1ccc(N2CCN(C3=C(Cl)C(=O)N(c4cccc(C(F)(F)F)c4)C3=O)CC2)cc1
\end{lstlisting}

\noindent\textbf{Model trajectory.}
The selected verbatim excerpt preserves the failed inventory query, the protecting-group decision,
the revised route, and the final verification; bracketed ellipses mark omitted intermediate analysis.
\ReasoningListing{Inventory-guided protecting-group revision}{case_studies/chembl_824_trace_excerpt.txt}

The first query identifies 1-(4-methoxyphenyl)piperazine as the sole unavailable leaf. In the revised
route, all three terminal precursors---N-Boc-piperazine, 4-bromoanisole, and the N-aryl
dichloromaleimide---are available, and the second query verifies the complete route.

\begin{figure}[H]
    \centering
    \includegraphics[width=0.88\linewidth]{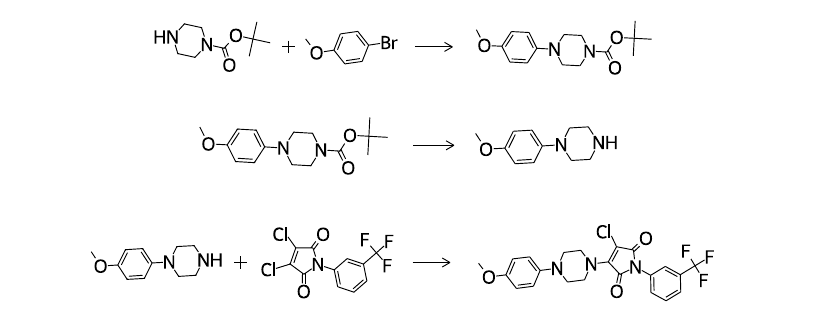}
    \caption{Inventory-verified protecting-group route rendered in forward synthetic order.}
    \label{fig:case-boc-route}
\end{figure}

\Needspace{8\baselineskip}
\ToolListing{Final answer}{case_studies/chembl_824_final_answer.json}

\noindent\textbf{Manual chemical review.}
We examined the route transformation by transformation in the forward direction. (1) Mono-Boc
piperazine exposes one nitrogen for palladium-catalyzed N-arylation with 4-bromoanisole while
blocking reaction at the second nitrogen. (2) Standard acidic Boc cleavage then reveals
1-(4-methoxyphenyl)piperazine without disturbing the anisole substituent. (3) The remaining
secondary amine can displace one vinylic chloride from the electron-poor N-aryl
3,4-dichloromaleimide; the two adjacent carbonyl groups activate this substitution, while the second
chloride is retained in the target. The sequence therefore assigns the Boc group a chemically
specific role: it controls which piperazine nitrogen is functionalized first and exposes the other
only when required for the final bond formation. The primary chemical-validity assessment accepts
all three transformations. As complementary orthogonal evidence, ChemCensor also supports every
step under both its reaction-center and functional-group criteria.

\noindent\textbf{Interpretation.}
The protecting group is introduced in direct response to an unavailable precursor rather than added
as an incidental decoration. Solvent preserves it through the selective arylation, removes it at the
appropriate point, and then uses the newly exposed nitrogen in the final convergent substitution. This
case provides a concrete example of inventory-guided revision yielding a chemically purposeful,
stock-complete synthesis strategy.

\subsection{Solvent Generates Diverse Plausible Routes for the Same Target}
\label{app:case-parallel-sampling}

\begin{figure}[H]
    \centering
    \includegraphics[width=0.42\linewidth]{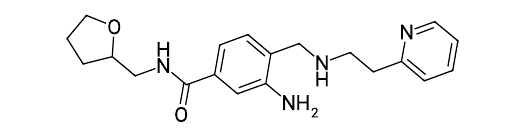}
    \caption{Target molecule shared across independently sampled trajectories.}
    \label{fig:case-parallel-target}
\end{figure}

The preceding examples show how Solvent revises a route within a single trajectory. Beyond this
iterative behavior, Solvent can propose different chemically plausible synthesis plans for the same
target. Figure~\ref{fig:case-parallel-target} presents stock-complete routes obtained in independent
trials that differ in both the reaction used to form the benzylic C--N bond and the available
aromatic precursor. We reproduce only the decisive verbatim excerpts; bracketed ellipses indicate
omitted reasoning.

\noindent\textbf{Target SMILES.}
\begin{lstlisting}[style=tooltrace]
Nc1cc(C(=O)NCC2CCCO2)ccc1CNCCc1ccccn1
\end{lstlisting}

\subsubsection{Route A: Benzyl Substitution}
\label{app:case-parallel-benzyl}

After an aldehyde-based proposal fails the inventory check, this trajectory changes the key
benzylic C--N disconnection from reductive amination to substitution with a benzyl bromide.

\ReasoningListing{Selected reasoning excerpt}{case_studies/parallel_144_benzyl_excerpt.txt}

\begin{figure}[H]
    \centering
    \includegraphics[width=0.88\linewidth]{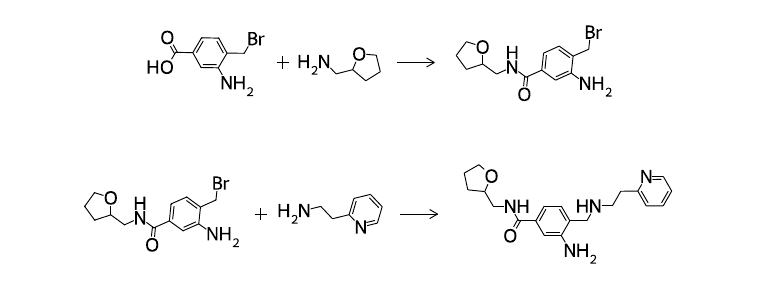}
    \caption{Stock-complete route based on benzyl substitution, rendered in forward synthetic order.}
    \label{fig:case-parallel-benzyl}
\end{figure}

The revised route first forms the aryl amide and then couples the resulting benzyl bromide with
2-(pyridin-2-yl)ethan-1-amine (Figure~\ref{fig:case-parallel-benzyl}). The subsequent inventory query
confirms that all three terminal precursors are available.

\subsubsection{Route B: Reductive Amination}
\label{app:case-parallel-reductive}

In another trial, Solvent retains the aldehyde disconnection but replaces the unavailable acid with
an available methyl ester, producing a different route to the same target.

\ReasoningListing{Selected reasoning excerpt}{case_studies/parallel_144_reductive_excerpt.txt}

\begin{figure}[H]
    \centering
    \includegraphics[width=0.88\linewidth]{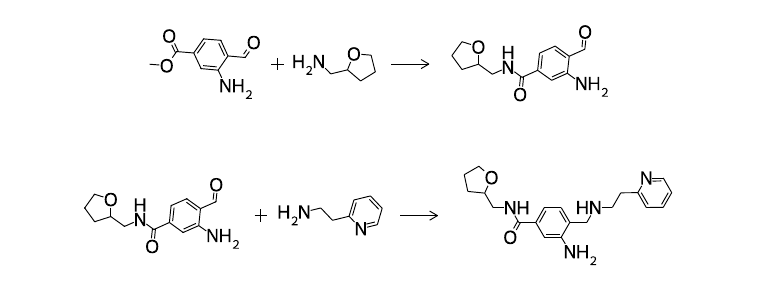}
    \caption{Stock-complete route centered on reductive amination, rendered in forward synthetic order.}
    \label{fig:case-parallel-reductive}
\end{figure}

In this proposal, aminolysis of the aromatic ester furnishes the amide-bearing aldehyde, which is
then coupled with 2-(pyridin-2-yl)ethan-1-amine by reductive amination
(Figure~\ref{fig:case-parallel-reductive}). The inventory query again confirms the availability of
all three terminal precursors.

\noindent\textbf{Interpretation.}
The routes construct the same benzylic C--N bond through different chemistry: one uses
nucleophilic substitution of a benzyl bromide, whereas the other uses reductive amination of an
aldehyde. They consequently rely on different aromatic precursors from the inventory. Together,
these trials show that Solvent can generate chemically plausible synthesis plans with distinct
reaction strategies for the same target.

\subsection{Representative Cases of Long-Horizon Reasoning across RL Stages}
\label{app:case-reasoning-evolution}

For each case, we compare outputs from policies before and after an RL stage on the same target.
The first comparison covers Round 1 RL and shows a shift from unstable teacher-guided traces to
concise and reliable interaction. The second covers final-stage RL and shows the post-RL policy
incorporating a protection-group strategy into the submitted route. These examples provide
qualitative views of how Solvent's long-horizon behavior changes after RL; bracketed ellipses mark
omitted text.

\subsubsection{From Unstable Imitation to Concise, Reliable Reasoning}
\label{app:case-round1-evolution}

Round 1 RL markedly improves the reliability of model-generated trajectories. Under matched
decoding, we compare 12,896 paired trajectories from the teacher-guided SFT policy and
$\pi_{\mathrm{seed}}$ (Table~\ref{tab:round1-rollout-health}). RL reduces truncation from 78.27\%
to 0.79\% and malformed or incomplete interactions from 71.32\% to 39.69\%, while increasing the
final-answer parse rate from 12.93\% to 70.91\%.

\begin{table}[H]
\centering
\caption{Rollout reliability before and after Round 1 RL under matched decoding. Malformed or
incomplete interactions fail the tool-trace validity check.}
\label{tab:round1-rollout-health}
\small
\setlength{\tabcolsep}{7pt}
\renewcommand{\arraystretch}{1.08}
\begin{tabularx}{\linewidth}{L{3.1cm}*{3}{Y}}
\toprule
Policy & \shortstack{Truncated\\(\%)} & \shortstack{Parseable final\\answer (\%)} & \shortstack{Malformed or\\incomplete interaction (\%)} \\
\midrule
Teacher-guided SFT & 78.27 & 12.93 & 71.32 \\
$\pi_{\mathrm{seed}}$ & 0.79 & 70.91 & 39.69 \\
\bottomrule
\end{tabularx}
\end{table}

A comparison on the same target illustrates this transition. The SFT policy generates 31,964
tokens and issues three unsuccessful inventory queries without producing a final answer. In
contrast, $\pi_{\mathrm{seed}}$ returns a stock-complete route after 2,856 tokens and a single
inventory query.

\ReasoningListing{Teacher-guided SFT: repeated speculative restarts}{case_studies/round1_sft_unstable_excerpt.txt}

\ReasoningListing{After Round 1 RL: route construction and verification}{case_studies/round1_rl_concise_excerpt.txt}

\noindent\textbf{Interpretation.}
The SFT trace cycles through increasingly speculative constructions without converging on an
executable route. In contrast, $\pi_{\mathrm{seed}}$ identifies two amide disconnections and submits
the resulting route for inventory verification. Once precursor availability is confirmed, the policy
returns the route without revisiting its plan. This case mirrors the aggregate transition in
Table~\ref{tab:round1-rollout-health} toward concise reasoning and reliable tool use.

\subsubsection{A Protection-Group Strategy after Final-Stage RL}
\label{app:case-final-rl-evolution}

We next compare two policies on the same temporary-protection target. Before final-stage RL, the
model identifies Boc protection as a way to control the free piperidine nitrogen but does not
incorporate it into the final route.
The submitted four-step route is stock-complete but fails the chemical-validity assessment.

\ReasoningListing{Round 2 bootstrapping: selected verbatim excerpt}{case_studies/bootstrapping_boc_excerpt.txt}

\begin{table}[H]
\centering
\caption{Protection-group strategy before and after final-stage RL.}
\label{tab:case-final-rl-evolution}
\small
\setlength{\tabcolsep}{7pt}
\renewcommand{\arraystretch}{1.08}
\begin{tabularx}{\linewidth}{L{3.2cm}*{4}{Y}}
\toprule
Policy & Tokens & Steps & \shortstack{Stock-\\complete} &
\shortstack{Chemical\\validity} \\
\midrule
Round 2 bootstrapping & 6,449 & 4 & Yes & No \\
Final-stage RL & 3,935 & 3 & Yes & Yes \\
\bottomrule
\end{tabularx}
\end{table}

\begin{figure}[!ht]
    \centering
    \includegraphics[width=0.78\linewidth]{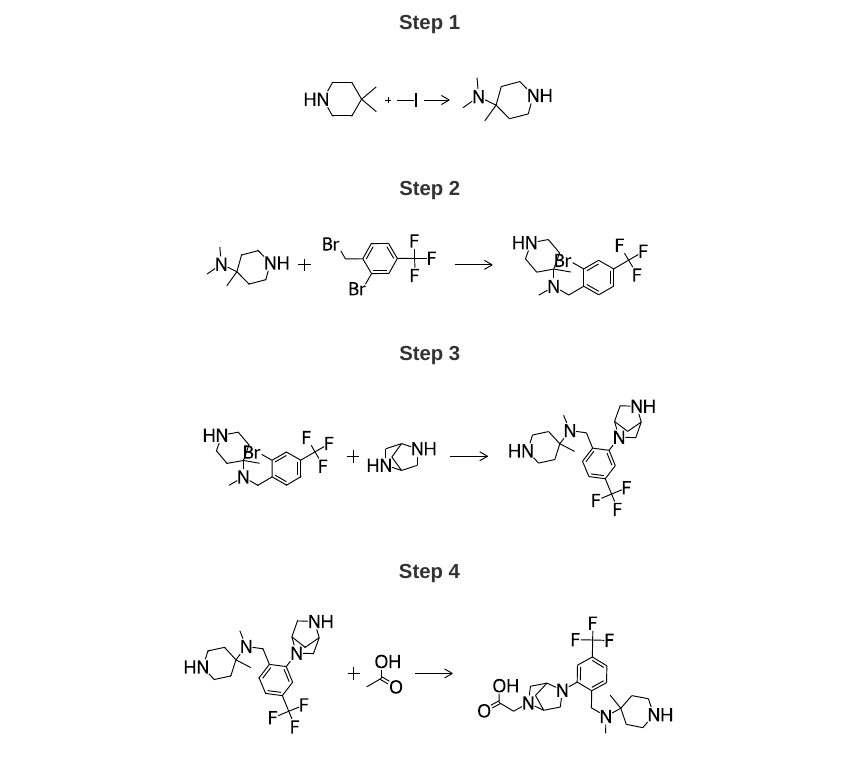}
    \par\vspace{-0.35em}
    {\small\textbf{(a) Round 2 bootstrapping}\par}
    \vspace{0.5em}
    \includegraphics[width=0.78\linewidth]{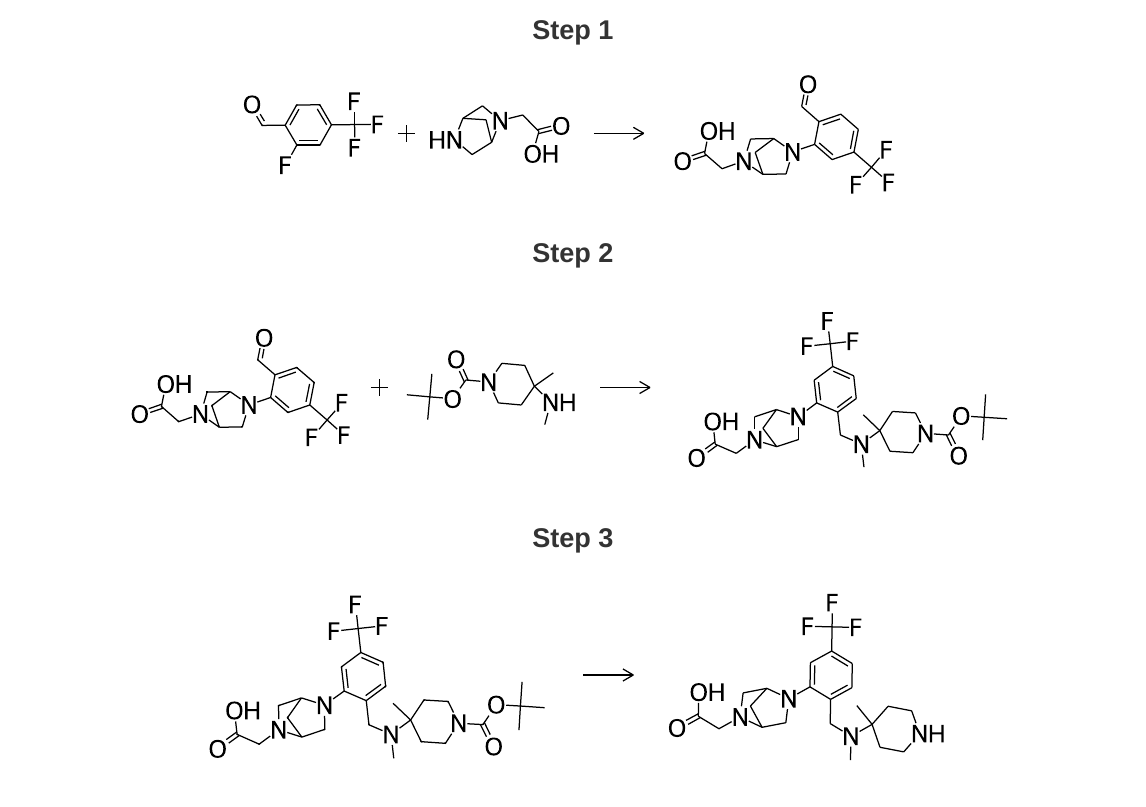}
    \par\vspace{-0.35em}
    {\small\textbf{(b) Final-stage RL}\par}
    \caption{Routes proposed before and after final-stage RL, with steps shown in forward synthetic
    order. (a) The Round 2 bootstrapping route is stock-complete but fails the chemical-validity
    check because its initial methylation does not account for the exocyclic nitrogen. (b) The
    final-stage RL route carries Boc protection through the selective bond formation and removes it
    only in the final step.}
    \label{fig:case-final-rl-evolution}
\end{figure}

After final-stage RL, the policy retains the Boc-protected piperidine through reductive amination
and removes the protecting group only after the selective bond formation. The complete final-stage
trajectory and route are shown in Figure~\ref{fig:case-final-rl-evolution}. The revised route
provides a concrete illustration of a more mature chemical strategy after
final-stage RL: temporary protection is not merely considered but incorporated at the appropriate
stage, enabling selective bond formation within a chemically valid route.

Figure~\ref{fig:case-final-rl-evolution} compares the two submitted routes.

\section{Further Discussion}
\label{app:limitations}

Framing multi-step retrosynthesis as a long-horizon task gives the LLM direct control over global
route planning and stepwise disconnection decisions, allowing the model to explore routes beyond
the candidates supplied by an external single-step model while making reliable route verification
essential, which however, remains a general open challenge in multi-step
retrosynthesis~\citep{Choe_2025}. Our post-generation evaluator combines a permissive
reaction-support check with atom-source consistency, providing a practical screen for evident
chemical inconsistencies while leaving route generation unrestricted.

Passing these checks and terminating in available building blocks, however, does not establish that
a route will succeed experimentally. Laboratory outcomes depend on reaction conditions,
chemo- and stereoselectivity, competing pathways, intermediate stability, purification,
reproducibility, and scale. The practical value of a route also reflects overall yield, material cost,
safety, and environmental impact~\citep{Genheden_2025}. These factors cannot be fully assessed
in silico and ultimately require experimental validation.

An important extension is to carry experimental observations across synthesis cycles and use them
in later route decisions. ACRA demonstrates that an LLM-based workflow can extract synthetic
procedures, translate them into executable robotic instructions, and verify them through real
experiments~\citep{Pagel2026}. Integrating this experimental interface into Solvent's autonomous
long-horizon planning loop would enable more direct evaluation of route outcomes, allow Solvent to
adapt its strategy to experimental feedback, and potentially raise its performance ceiling while
increasing its practical utility.

\end{document}